\documentclass{article} %
\pdfoutput=1 %
\usepackage{nips15submit_e,times}
\usepackage{hyperref}
\usepackage{url}
\usepackage{amsthm,amsmath}
\usepackage{chngcntr}

\newtheorem{lemma}{Lemma}
\newtheorem{theorem}{Theorem}
\newtheorem{corollary}{Corollary}
\newtheorem{definition}{Definition}

\newcommand{\Prob}{{{\bf P}}}

\renewcommand{\L}{{\mathcal L}}

\DeclareMathOperator*{\argmin}{argmin}

\usepackage{times}
\usepackage{graphicx} %
\usepackage{subfigure} 

\usepackage{algorithm}
\usepackage{algorithmic}

\usepackage{amssymb, amsmath}
\usepackage{amsthm}
\usepackage{booktabs}  %
\usepackage{color}
\usepackage{centernot}
\usepackage{mathtools}
\usepackage{multirow}
\usepackage{sidecap}
\usepackage{tabularx}

\newcommand{\E}{{\mathbf E}}
\newcommand{\Efull}{\mathbb{E}}

\newcolumntype{S}{>{\centering\arraybackslash} m{.10\linewidth} }
\newcolumntype{T}{>{\centering\arraybackslash} m{.28\linewidth} }
\newcolumntype{U}{>{\centering\arraybackslash} m{.08\linewidth} }
\newcolumntype{V}{>{\centering\arraybackslash} m{.37\linewidth} }
\newcolumntype{W}{>{\centering\arraybackslash} m{.1\linewidth} }

\graphicspath{{images/}{figures/}}   %

\title{Variance Reduced Stochastic Gradient Descent \\with Neighbors}

\author{
Thomas Hofmann\\
Department of Computer Science\\
ETH Zurich, Switzerland
\And
Aurelien Lucchi\\
Department of Computer Science\\
ETH Zurich, Switzerland
\And
Simon Lacoste-Julien\\
INRIA - Sierra Project-Team\\
\'{E}cole Normale Sup\'{e}rieure, Paris, France 
\And
Brian McWilliams\\
Department of Computer Science\\
ETH Zurich, Switzerland
}

\nipsfinalcopy %

\begin{document}

\maketitle

\begin{abstract} 

Stochastic Gradient Descent (SGD) is a workhorse in machine learning, yet its slow convergence can be a computational bottleneck. Variance reduction techniques such as SAG, SVRG and SAGA have been proposed to overcome this weakness, achieving linear convergence. However, these methods are either based on computations of full gradients at pivot points, or on keeping per data point corrections in memory. Therefore speed-ups relative to SGD may need a minimal number of epochs in order to materialize. This paper investigates algorithms that can exploit neighborhood structure in the training data to share and re-use information about past stochastic gradients across data points, which offers advantages in the transient optimization phase.  As a side-product we provide a unified convergence analysis for a family of variance reduction algorithms, which we call memorization algorithms. We provide experimental results supporting our theory. 
\end{abstract}

\section{Introduction}

We consider a general problem that is pervasive in machine learning, namely optimization of an empirical or regularized convex risk function. Given a convex loss $l$ and a $\mu$-strongly convex regularizer $\Omega$, one aims at finding a parameter vector $w$ which minimizes the (empirical) expectation:
\begin{align}
w^* = \argmin_w f(w), \quad f(w) = \frac 1n \sum_{i=1}^n f_i(w), \quad f_i(w) := l(w, (x_i,y_i)) + \Omega(w)\,.
\label{eq:problem}
\end{align}
We assume throughout that each $f_i$ has $L$-Lipschitz-continuous gradients. Steepest descent can find the minimizer $w^*$, but requires repeated computations of full gradients $f'(w)$, which becomes prohibitive for massive data sets. Stochastic gradient descent (SGD) is a popular alternative, in particular in the context of large-scale learning~\cite{bottou2010, shalev2011}. SGD updates only involve $f'_i(w)$ for an index $i$ chosen uniformly at random, providing an unbiased gradient estimate, since $\E f'_i(w) = f'(w)$. 

It is a surprising recent finding \cite{shalev2013stochastic,johnson2013,schmidt2013minimizing,konevcny2013} that the finite sum structure of $f$ allows for significantly faster convergence in expectation. Instead of the standard $O(1/t)$ rate of SGD for strongly-convex functions, it is possible to obtain linear convergence with geometric rates. While SGD requires asymptotically vanishing learning rates, often chosen to be $O(1/t)$ \cite{robbins1951}, these more recent methods introduce  corrections that ensure convergence for constant learning rates. 

Based on the work mentioned above, the contributions of our  paper are as follows: First, we define a family of variance reducing SGD algorithms, called memorization algorithms, which includes  SAGA and SVRG as special cases, and develop a unifying  analysis technique for it. Second, we show geometric rates for all step sizes $\gamma < \frac 1{4L}$, including a universal ($\mu$-independent) step size choice, providing the first $\mu$-adaptive convergence proof for SVRG. Third, based on the above analysis, we present new insights into the trade-offs between freshness and biasedness of the corrections computed from previous stochastic gradients. Fourth, we propose a new class of algorithms that resolves this trade-off by computing corrections based on stochastic gradients at neighboring points. We experimentally show its benefits  in the regime of  learning with  a small number of epochs.

\section{Memorization Algorithms}
\label{sect:memorization}

\subsection{Algorithms} 

\paragraph*{Variance Reduced SGD}

Given an optimization problem as in \eqref{eq:problem}, we investigate a class of stochastic gradient descent algorithms that generates an iterate sequence $w^t$ ($t \ge 0$) with updates taking the  form:
\begin{align}
w^+ = w - \gamma g_i(w),  \quad g_i(w)=  f'_i(w) - \bar \alpha_i \quad \text{with} \quad  \bar{\alpha}_i := \alpha_i - \bar{\alpha},
\label{eq:sgd-corrected}
\end{align}
where $\bar{\alpha} := \frac 1n \sum_{j=1}^n \alpha_j$.
Here $w$ is the current and $w^+$ the new parameter vector, $\gamma$ is the  step size,  and $i$ is an index selected uniformly at random. $\bar\alpha_i$ are variance correction terms such that  $\E \bar\alpha_i = 0$, which guarantees unbiasedness $\E g_i(w) =f'(w)$. The aim is to define updates of asymptotically vanishing variance, i.e.~$g_i(w) \to 0$ as $w \to w^*$, which  requires  $\bar\alpha_i \to f'_i(w^*)$. This implies that corrections need to be designed in a way to exactly cancel out the stochasticity of $f'_i(w^*)$ at the optimum. How the \emph{memory} $\alpha_j$ is updated distinguishes the different algorithms that we consider. 

\paragraph*{SAGA}

The SAGA algorithm \cite{defazio2014} maintains variance corrections $\alpha_i$  by memorizing stochastic gradients. The update rule is $\alpha^+_i  = f'_i(w)$ for the selected $i$, and $\alpha_j^+= \alpha_j$, for $j \neq i$. Note that these corrections will be used the next time the same index $i$ gets sampled.  Setting $\bar\alpha_i := \alpha_i - \bar \alpha$ guarantees unbiasedness. Obviously, $\bar \alpha$ can be updated incrementally. SAGA reuses the stochastic gradient $f_i'(w)$ computed at step $t$ to update $w$ as well as $\bar \alpha_i$.

\paragraph*{$q$-SAGA}
We  also consider $q$-SAGA, a method that updates $q \ge 1$ randomly chosen $\alpha_j$ variables at each iteration. This is a convenient reference point to investigate the advantages of ``fresher" corrections. Note that in SAGA the corrections will be on average $n$ iterations ``old". In $q$-SAGA this can be controlled to be $n/q$ at the expense of additional gradient computations.

\paragraph*{SVRG} 

We reformulate a variant of SVRG \cite{johnson2013} in our framework using a randomization argument similar to (but simpler than) the one suggested in \cite{konevcny2013}. Fix $q>0$ and draw in each  iteration $r \sim \text{Uniform}[0;1)$. If $ r< q/n$, a complete update, $\alpha_j^+  =  f_j'(w)$ ($\forall j$) is performed, otherwise they are left unchanged. While $q$-SAGA updates exactly $q$ variables in each iteration, SVRG occasionally updates all $\alpha$ variables by triggering an additional sweep through the data. There is an option to not maintain $\alpha$ variables explicitly and to save on space  by storing only $\bar \alpha = f'(w)$ and $w$.

\paragraph*{Uniform Memorization Algorithms} 

Motivated by SAGA and SVRG, we define a class of algorithms, which we call \textit{uniform memorization algorithms}. 
\begin{definition}
\label{def:memorization} 
A uniform $q$-memorization algorithm evolves iterates $w$ according to Eq.~\eqref{eq:sgd-corrected} and selects in each iteration a random index set $J$ of memory locations to update according to
\begin{align}
\alpha_j^+ := 
\begin{cases} 
f'_j(w) & \text{if $j \in J$} \\
\alpha_j & \text{otherwise,}
\end{cases}
\label{eq:memorization}
\end{align}
such that any $j$ has the \emph{same} probability of $q/n$ of being updated, i.e. $\forall j$, $\sum_{J \ni j} \Prob\{ J \} = \frac {q}{n}$.
\end{definition}
Note that $q$-SAGA  and the above  SVRG are special cases. For $q$-SAGA: $\Prob\{J\}=1/\binom n q$ if $|J|=q$ $\Prob\{J\}=0$ otherwise. For SVRG: $\Prob\{ \emptyset\} = 1-q/n$, $\Prob\{ [1:n]\}=q/n$, $\Prob\{J\}=0$, otherwise. 

\paragraph{${\mathcal N}$-SAGA}
Because we need it in Section \ref{sect:sharing}, we will also define an algorithm, which we call ${\cal N}$-SAGA, which makes use of a neighborhood system ${\cal N}_i \subseteq \{1,\dots,n\}$ and which selects neighborhoods uniformly, i.e.~$\Prob\{{\cal N}_i\}= \frac 1n$.  Note that Definition \ref{def:memorization} requires $| \{i: j \in {\cal N}_i\}|=q$ $(\forall j)$. 

Finally, note that for generalized linear models where $f_i$ depends on $x_i$ only through $\langle w, x_i \rangle$, we get $f_i'(w) = \xi'_i(w) x_i$, i.e.~the update direction is determined by $x_i$, whereas the effective step length depends on the derivative of a scalar function $\xi_i(w)$. As used in~\cite{schmidt2013minimizing}, this leads to significant memory savings as one only needs to store the scalars $\xi'_i(w)$ as $x_i$ is always \textit{given} when performing an update. 

\subsection{Analysis}
\paragraph*{Recurrence of Iterates}

The evolution equation \eqref{eq:sgd-corrected} in expectation implies the recurrence (by crucially using the unbiasedness condition $\E g_i(w) =f'(w)$):
\begin{align}
\E \| w^+\! -\! w^* \|^2 & %
= \| w - w^* \|^2 - 2 \gamma \langle f'(w), w-w^*\rangle + \gamma^2 \E \| g_i(w) \|^2 \,.
\label{eq:recurrence}
\end{align}
Here and in the rest of this paper, expectations are always taken only with respect to $i$ (conditioned on the past).   We utilize a number of bounds (see \cite{defazio2014}), which exploit strong convexity of $f$ (wherever $\mu$ appears) as well as Lipschitz continuity of the $f_i$-gradients (wherever $L$ appears):
\begin{align}
 \langle f'(w), w-w^* \rangle  & \ge  f(w) - f(w^*)+ \tfrac \mu 2  \| w-w^*\|^2\,,
\label{eq:naive-strong-convexity} \\
\E \|g_i(w)\|^2 & \le 2  \E \| f'_i(w) - f'_i(w^*) \|^2 + 2 \E \| \bar \alpha_i - f_i'(w^* )\|^2 \,,
\label{eq:beta-split} 
\\
\| f'_i(w) - f_i'(w^*) \|^2 & \le 2L h_i(w), \quad h_i(w)  := f_i(w) - f_i(w^*)-  \langle w-w^*, f_i'(w^*) \rangle
\label{eq:smoothness-bound}
\,, \\
\! \E \| f_i'(w) \!-\! f_i'(w^*)\|^2 & \le 2L f^\delta(w), \quad f^\delta(w) := f(w) - f(w^*) \,, 
\label{eq:expected-smoothness}
\\
\E \| \bar \alpha_i - f_i'(w^*)\|^2 & = \E \| \alpha_i - f_i'(w^*) \|^2 - \| \bar \alpha\|^2 \le \E \| \alpha_i - f_i'(w^*) \|^2 .
\label{eq:alpha-simplification}
\end{align}
Eq.~\eqref{eq:beta-split} can be generalized~\cite{defazio2014} using $\| x \pm y\|^2 \le (1+\beta) \| x\|^2 +  (1+\beta^{-1}) \|y\|^2$ with $\beta>0$. However for the sake of simplicity, we sacrifice tightness and choose $\beta=1$.
Applying all of the above yields:
\begin{lemma} 
\label{lemma:w-recurrence}
For the iterate sequence of any algorithm that evolves solutions according to Eq.~\eqref{eq:sgd-corrected}, the following holds for a single update step, in expectation over the choice of $i$:
\begin{equation*}
\| w - w^*\|^2  - \E \| w^+ - w^*\|^2 \ge \,\, \gamma \mu \| w- w^*\|^2  - 2 \gamma^2 \E \| \alpha_i - f_i'(w^*)\|^2  + \left( 2\gamma - 4\gamma^2 L \right) f^\delta(w) \,.
\end{equation*}
\end{lemma}
All proofs are deferred to the Appendix. 

\paragraph*{Ideal and Approximate Variance Correction}
 
Note that in the ideal case of $\alpha_i = f'_i(w^*)$, we would immediately get a condition for a contraction by choosing $\gamma = \frac{1}{2L}$, yielding a  rate of $1- \rho$ with $\rho = \gamma \mu = \frac{\mu}{2L}$, which is half the inverse of the condition number $\kappa:=L/\mu$.

How can we further bound  $\E \| \alpha_i - f_i'(w^*) \|^2$ in the case of ``non-ideal" variance-reducing SGD? A key insight is that for  memorization algorithms, we can apply the smoothness bound in Eq.~\eqref{eq:smoothness-bound}
\begin{align}
\| \alpha_i-f'_i(w^*)\|^2 
= \| f'_i(w^{\tau_i}) -f'_i(w^*)\|^2 
\le 2L  h_i(w^{\tau_i}), \quad \text{(where $w^{\tau_i}$ is old $w$)} \,.
\label{eq:alpha-bound}
\end{align}
Note that if we only had approximations $\beta_i$ in the sense that $\| \beta_i - \alpha_i \|^2 \le \epsilon_i$  (see Section \ref{sect:sharing}), then we can use $\|x-y\| \leq 2 \|x\| + 2 \|y\|$ to get the somewhat worse bound:
\begin{align}
\| \beta_i-f'_i(w^*)\|^2  \le 2\| \alpha_i-f'_i(w^*)\|^2  + 2 \| \beta_i - \alpha_i\|^2
\le 4L  h_i(w^{\tau_i})  + 2 \epsilon_i.
\label{eq:alpha-tilde-bound}
\end{align}

\paragraph*{Lyapunov Function}

Ideally, we would like to show that for a suitable choice of $\gamma$, each iteration results in a contraction  $\E \| w^+ - w^*\|^2 \le (1-\rho) \| w - w^*\|^2$, where $0 < \rho \leq 1$. However, the main challenge arises from the fact that the quantities $\alpha_i$ represent stochastic gradients from previous iterations. This requires a somewhat more complex proof technique. Adapting the Lyapunov function method from \cite{defazio2014}, we  define upper bounds $H_i \geq \| \alpha_i - f'_i(w^*)\|^2$ such that $H_i \to 0$ as $w \to w^*$. We start with $\alpha^0_i\!=\!0$ and (conceptually) initialize $H_i = \|f'_i(w^*)\|^2$, and then update $H_i$ in sync with $\alpha_i$,
\begin{align}
H_i^+ := 
\begin{cases}
2L \, h_i(w) & \text{if $\alpha_i$ is updated} \\
H_i & \text{otherwise}
\end{cases} 
\label{eq:recurrence-h1}
\end{align} 
so that we always maintain valid bounds $\| \alpha_i - f_i'(w^*) \|^2 \le H_i$ and $\E\| \alpha_i - f_i'(w^*) \|^2 \le \bar H$ with $\bar H := \frac 1n \sum_{i=1}^n H_i$. The $H_i$ are quantities showing up in the analysis, but need  \textit{not} be computed. We now define a $\sigma$-parameterized family of Lyapunov functions\footnote{This is a simplified version of the one appearing in~\cite{defazio2014}, as we assume $f'(w^*)=0$ (unconstrained regime).}
\begin{align} \label{eq:lyapunov_fct}
\L_\sigma(w,H) := \| w- w^*\|^2 + S \sigma \,  \bar H, 
\quad \text{with} \; \; S:= \left( \frac{\gamma n}{L q} \right) \quad \text{and}\quad  0\leq  \sigma \leq 1\,.
\end{align}
In expectation under a random update, the Lyapunov function $\L_\sigma$ changes as $ \E \L_\sigma(w^+,H^+)  = \E \| w^+ - w^*\|^2  + S \sigma\, \E \bar H^+$. 
We can readily apply Lemma \ref{lemma:w-recurrence} to bound the first part. The second part is due to \eqref{eq:recurrence-h1}, which mirrors the update of the $\alpha$ variables.  
By crucially using the property that any $\alpha_j$ has the same probability of being updated in~\eqref{eq:memorization}, we
get the following result: 
\begin{lemma} 
For a uniform $q$-memorization algorithm, it holds that 
\begin{align}
\E \bar H^+ = \left( \frac{n-q}{n} \right) \bar H + \frac{2Lq}{n} \, f^\delta(w) .
\label{eq:h-recurrence}
\end{align}
\label{lemma:hrecurrence2}
\end{lemma}
\vspace{-3.5mm}
Note that in expectation the shrinkage does not depend on the location of previous iterates $w^\tau$ and the new increment is  proportional to the sub-optimality of the current iterate $w$.  Technically, this is how the possibly complicated dependency on previous iterates is dealt with in an effective manner. 

\paragraph*{Convergence Analysis}

We first state our main Lemma about Lyapunov function contractions:
\begin{lemma}
\label{lemma:lyapunov}
Fix $c \in (0;1]$ and $\sigma \in [0;1]$ arbitrarily. For any uniform $q$-memorization algorithm with sufficiently small step size $\gamma$ such that
\begin{align}
\gamma \le \frac{1}{2L} \min \left\{\frac{K\sigma }{K + 2c\sigma}, 1-\sigma \right\}, \quad 
\text{and} \quad K:=  \frac{4qL}{n \mu},
\label{eq:gamma-admissible}
\end{align}
we have that 
\begin{align}
\E \L_\sigma(w^+,H^+)  \le (1-\rho) \L_\sigma(w,H),\quad \text{with} \quad \rho := c \mu \gamma.
\end{align}
Note that $\gamma < \frac 1{2L} \max_{\sigma \in [0,1]} \min\{ \sigma ,  1-\sigma\} = \frac{1}{4L}$ (in the $c \to 0$ limit).
\end{lemma}
By maximizing the bounds in Lemma \ref{lemma:lyapunov} over the choices of $c$ and $\sigma$, we obtain our main result that provides guaranteed geometric rates for all step sizes up to $\frac 1 {4L}$. 
\begin{theorem}
\label{theorem:main}
Consider  a uniform $q$-memorization algorithm. For any step size $\gamma = \frac a {4L}$ with $a<1$, the algorithm converges at a geometric rate of at least $(1-\rho(\gamma))$ with 
\begin{align}
\rho(\gamma) = \frac{q}{n} \cdot \frac{1 - a}{1-a/2} 
= \frac{\mu}{4L} \cdot \frac{K (1 - a)}{1-a/2}, \;\;  \text{if} \;\gamma \geq \gamma^*(K), \;\; 
\text{otherwise} \;\; \rho(\gamma) = \mu \gamma   \;\; 
\end{align}
where 
\begin{align}
\gamma^*(K) :=  \frac {a^*(K)} {4 L}, \quad 
a^*(K) := \frac{2K}{1+K + \sqrt{1+K^2}},
\quad K:=  \frac{4qL}{n \mu} = \frac{4q}{n} \kappa \, .
\label{eq:gammalk}
\end{align}
\end{theorem}
We would like to provide more insights into this result. 
\begin{corollary}
\label{corollary:opt-rho}
In Theorem \ref{theorem:main}, $\rho$ is maximized for $\gamma = \gamma^*(K)$. We can write $\rho^*(K) = \rho(\gamma^*)$ as 
\begin{align}
\rho^*(K) = 
\frac \mu {4L} a^*(K) %
= \frac{q}{n}\frac{a^*(K)}{K} = \frac qn \left[ \frac{2}{1+K + \sqrt{1 + K^2} } \right]
\end{align}
In the big data regime $\rho^* = \frac qn (1-\frac 12 K + O(K^3))$, whereas in the ill-conditioned case $\rho^* = \frac \mu{4L} (1 - \frac {1}{2}K^{-1} + O(K^{-3}))$.
\end{corollary}
The guaranteed rate is bounded by $\frac{\mu}{4L}$ in the regime where the condition number dominates $n$  (large $K$)  and by $\frac qn$ in the opposite regime of large data (small $K$). Note that if $K \le 1$, we have $\rho^* = \zeta  \frac qn$ with  $\zeta \in [2/(2 + \sqrt{2}); 1]  \approx [0.585;1]$. So for $q \le n \frac{\mu}{4L}$, it pays off to increase freshness as it affects the rate proportionally.  In the ill-conditioned regime ($\kappa > n$), the influence of $q$ vanishes. 

Note that for $\gamma \ge \gamma^*(K)$, $ \gamma \to \frac1{4L}$ the rate decreases monotonically, yet the decrease is only minor. With the exception of a small neighborhood around $\frac{1}{4L}$, the entire range of $\gamma \in [\gamma^*; \frac{1}{4L})$ results in very similar rates. Underestimating $\gamma^*$  however leads to a (significant) slow-down by a factor $\gamma/\gamma^*$.

As the optimal choice of $\gamma$ depends on $K$, i.e.~$\mu$, we would prefer step sizes that are $\mu$-independent, thus giving rates that adapt to the local curvature  (see \cite{schmidt2013minimizing}). It turns out that by choosing a step size that maximizes $\min_K \rho(\gamma)/\rho^*(K)$, we obtain a $K$-agnostic step size with rate off by at most $1/2$:
\begin{corollary}
\label{corollary:universal}
Choosing $\gamma = \frac{2 -\sqrt{2} }{4L}$, leads to $\rho(\gamma) \ge (2 - \sqrt{2}) \rho^*(K) > \frac 12 \rho^*(K)$ for all $K$.
\end{corollary}
To gain more insights into the trade-offs for these fixed large universal step sizes, the following corollary details the range of rates obtained:
\begin{corollary}
Choosing $\gamma = \frac{a}{4L}$ with $ a<1$ yields $\rho = \min\{ \frac {1-a}{1 - \frac 12 a} \frac qn, \frac{a}{4} \frac \mu L \}$. In particular, we have for the choice $\gamma = \frac{1}{5L}$ that $\rho = \min\{ \frac {1}{3} \frac qn, \frac 15 \frac \mu L \}$ (roughly matching the rate given in~\cite{defazio2014} for $q=1$).
\label{corollary:universal2}
\end{corollary}

\section{Sharing Gradient Memory}
\label{sect:sharing}

\subsection{$\epsilon$-Approximation Analysis} 

As we have seen, fresher gradient memory, i.e.~a larger choice for $q$, affects the guaranteed convergence rate as $\rho \sim q/n$. However, as long as one step of a $q$-memorization algorithm is as expensive as $q$ steps of a $1$-memorization algorithm, this insight does not lead to practical improvements \textit{per se}. Yet, it raises the question, whether we can accelerate these methods, in particular ${\cal N}$-SAGA, by approximating gradients stored in the $\alpha_i$ variables. Note that we are always using the correct stochastic gradients in the \textit{current} update and by assuring $\sum_{i} \bar\alpha_i=0$, we will not introduce any bias in the update direction. Rather, we lose the guarantee of asymptotically vanishing variance at $w^*$. However, as we will show, it is possible to retain geometric rates up to a $\delta$-ball around $w^*$. 

We will focus on SAGA-style updates for concreteness and investigate an algorithm that mirrors ${\cal N}$-SAGA with the only difference that it maintains approximations $\beta_i$ to the true $\alpha_i$ variables. We aim to guarantee  $\E\| \alpha_i - \beta_i \|^2 \le \epsilon$ and will use Eq.~\eqref{eq:alpha-tilde-bound} to modify the right-hand-side of Lemma \ref{lemma:w-recurrence}. We see that approximation errors $\epsilon_i$ are multiplied with $\gamma^2$, which implies that we should aim for small learning rates, ideally without compromising the ${\cal N}$-SAGA rate. From Theorem \ref{theorem:main} and Corollary \ref{corollary:opt-rho} we can see that we can choose $\gamma \lesssim  q/{\mu n}$ for $n$ sufficiently large, which indicates that there is hope to dampen the effects of the approximations. We now make this argument more precise.
\begin{theorem}
\label{theorem:approximate}
Consider a uniform $q$-memorization algorithm with $\alpha$-updates that are on average $\epsilon$-accurate (i.e. $\E\| \alpha_i - \beta_i \|^2 \le \epsilon$). For any step size $\gamma \le \tilde{\gamma}(K)$, where $\tilde{\gamma}$ is given by Corollary \ref{corollary:patch} in the appendix (note that $\tilde{\gamma}(K) \geq \frac{2}{3} \gamma^*(K)$ and $\tilde{\gamma}(K) \to \gamma^*(K)$ as $K \to 0$), we get 
\begin{align}
\Efull \L(w^t,H^t) \le  (1-\mu \gamma)^t  \L_0+ \frac{4 \gamma \epsilon}{\mu}, \quad \text{ with } \L_0 := \|w^0 - w^*\|^2 + s(\gamma) \E \|f_i(w^*)\|^2 ,
\end{align}
where $\Efull$ denote the (unconditional) expectation over histories (in contrast to $\E$ which is conditional), and $ s(\gamma) := \frac{4 \gamma}{K \mu} (1-2 L \gamma)$.
\end{theorem} 
\begin{corollary}
With $\gamma = \min\{\mu , \tilde{\gamma}(K)\}$ we have 
\begin{align}
\frac {4 \gamma \epsilon }{\mu}  \le 4 \epsilon, \qquad 
\text{with a rate} \quad \rho = \min\{\mu^2, \mu \tilde{\gamma}\} \, .
\end{align}
\end{corollary} 
\vspace{-1.5mm}
In the relevant case of $\mu \sim 1/\sqrt {n}$, we thus converge towards some $\sqrt{\epsilon}$-ball around $w^*$ at a similar rate as for the exact method. For $\mu \sim n^{-1}$, we have to reduce the step size significantly to compensate the extra variance and to still converge to an $\sqrt{\epsilon}$-ball, resulting in the slower rate $\rho \sim n^{-2}$, instead of $\rho \sim n^{-1}$.

We also note that the geometric convergence of SGD with a constant step size to a neighborhood of the solution (also proven in~\cite{schmidt2014convergence}) can arise as a special case in our analysis. By setting $\alpha_i = 0$ in Lemma~\ref{lemma:w-recurrence}, we can take $\epsilon = \E \|f'_i(w^*)\|^2$ for SGD. An approximate $q$-memorization algorithm can thus be interpreted as making $\epsilon$ an algorithmic parameter, rather than a fixed value as in SGD.

\subsection{Algorithms} 

\paragraph*{Sharing Gradient Memory} 

We now discuss our proposal of using neighborhoods for sharing gradient information between close-by data points. Thereby we avoid an increase in gradient computations relative to $q$- or ${\cal N}$-SAGA at the expense of suffering an approximation bias. This leads to a new tradeoff between freshness  and approximation quality, which can be resolved in non-trivial ways, depending on the desired final optimization accuracy. 

We distinguish two types of quantities. First, the gradient memory $\alpha_i$ as defined by the reference algorithm ${\cal N}$-SAGA. Second, the shared gradient memory state $\beta_i$, which is used in a modified update rule in Eq.~\eqref{eq:sgd-corrected}, i.e.~$w^+ = w - \gamma ( f'_i(w) - \beta_i + \bar \beta )$. Assume that we select an index $i$ for the weight update, then we generalize Eq.~\eqref{eq:memorization} as follows
\begin{align}
\beta^+_j := \begin{cases} 
f'_i(w) & \text{if $j \in {\cal N}_i$} \\
\beta_j & \text{otherwise}
\end{cases}, \qquad \bar \beta := \frac 1n \sum_{i=1}^n \beta_i, \quad \bar \beta_i := \beta_i - \bar \beta \,.
\label{eq:shared-memorization}
\end{align}

In the important case of generalized linear models, where  one has 
$f'_i(w) = \xi'_i(w) x_i$, we can  modify the relevant case in Eq.~\eqref{eq:shared-memorization} by $\beta_j^+ := \xi'_i(w) x_j$. This has the advantages of using the correct direction, while reducing storage requirements. 

\paragraph*{Approximation Bounds}

For our analysis, we need to control the error $\| \alpha_i - \beta_i \|^2 \le \epsilon_i$. This obviously requires problem-specific investigations. 

Let us first look at the case of ridge regression. $f_i(w) := \frac 12 (\langle x_i, w \rangle - y_i)^2 + \frac \lambda 2 \| w\|^2$ and thus $f'_i(w) = \xi'_i(w) x_i + \lambda w$ with $\xi'_i(w) := \langle x_i, w \rangle - y_i$. Considering $j \in {\cal N}_i$ being updated, we have
\begin{align}
\| \alpha_j^+ - \beta_j^+\| = | \xi'_j(w) - \xi'_i(w) | \| x_j\|
\le \left( \delta_{ij}  \|w\| +|y_j - y_i| \right) \| x_j\| =: \epsilon_{ij}(w)
\label{eq:bound-quadratic}
\end{align}
where $\delta_{ij} := \| x_i - x_j \|$. Note that this can be pre-computed with the exception of the norm $\| w\|$ that we only know at the time of an update. 

Similarly, for regularized logistic regression with $y \in \{-1,1\}$, we have $\xi'_i(w) = y_i/(1 + e^{y_i \langle x_i, w\rangle})$. With the requirement on neighbors that $y_i = y_j$ we get 
\begin{align}
\| \alpha^+_j - \beta^+_j\| \le \frac{e^{\delta_{ij} \| w\|}-1}{1+ e^{-\langle x_i, w \rangle}} \|x_j\|
 =: \epsilon_{ij}(w)
\label{eq:bound-logistic}
\end{align}
Again, we can pre-compute $\delta_{ij}$ and $\| x_j\|$. In addition to $\xi'_i(w)$ we can also store $\langle x_i, w \rangle$. %

\paragraph*{$\epsilon {\cal N}$-SAGA}

We can use these bounds in two ways. First, assuming that the iterates stay within a norm-ball (e.g.~$L_2$-ball), we can derive upper bounds
\begin{align}
\epsilon_j(r) \ge  \max\{ \epsilon_{ij}(w): j \in {\cal N}_i, \; \|w\| \le r\}, \qquad \epsilon(r) = \frac 1n \sum_{j} \epsilon_j(r)\,. 
\end{align}
Obviously, the more compact the neighborhoods are, the smaller $\epsilon(r)$. This is most useful for the analysis. 
 Second, we can specify a target accuracy $\epsilon$ and then prune neighborhoods dynamically. This approach is more practically relevant as it allows us to directly control $\epsilon$. However, a dynamically varying neighborhood violates Definition \ref{def:memorization}. We fix this in a sound manner by modifying the memory updates as follows:
 \begin{align}
\beta^+_j := \begin{cases} 
f'_i(w) & \text{if $j \in  {\cal N}_i$ and $\epsilon_{ij}(w) \le \epsilon$} \\
f'_j(w) & \text{if $j \in  {\cal N}_i$ and $\epsilon_{ij}(w) > \epsilon$} \\
\beta_j & \text{otherwise}
\end{cases}
\label{eq:en-saga}
 \end{align}
This allows us to interpolate between sharing more aggressively (saving computation) and performing more computations in an exact manner. In the limit of $\epsilon \to 0$, we  recover ${\cal N}$-SAGA, as $\epsilon \to \epsilon^{\max}$ we recover the first variant mentioned. 

\paragraph*{Computing Neighborhoods}

Note that the pairwise Euclidean distances show up in the bounds in Eq.~\eqref{eq:bound-quadratic} and \eqref{eq:bound-logistic}. In the classification case we also require $y_i = y_j$, whereas in the ridge regression case, we also want $| y_i - y_j|$ to be small. Thus modulo filtering, this suggests the use of Euclidean distances as the metric for defining neighborhoods. Standard approximation techniques for finding near(est) neighbors can be used. This comes with a computational overhead, yet the additional costs will amortize over multiple runs or multiple data analysis tasks.

\begin{figure*}
  \begin{center}
    \begin{tabular}{@{}c@{\hspace{2mm}}c@{\hspace{2mm}}c@{}}
      (a) {\it Cov} &
      (b) {\it Ijcnn1} &
      (c) {\it Year} \\    
      \includegraphics[width=0.32\linewidth]{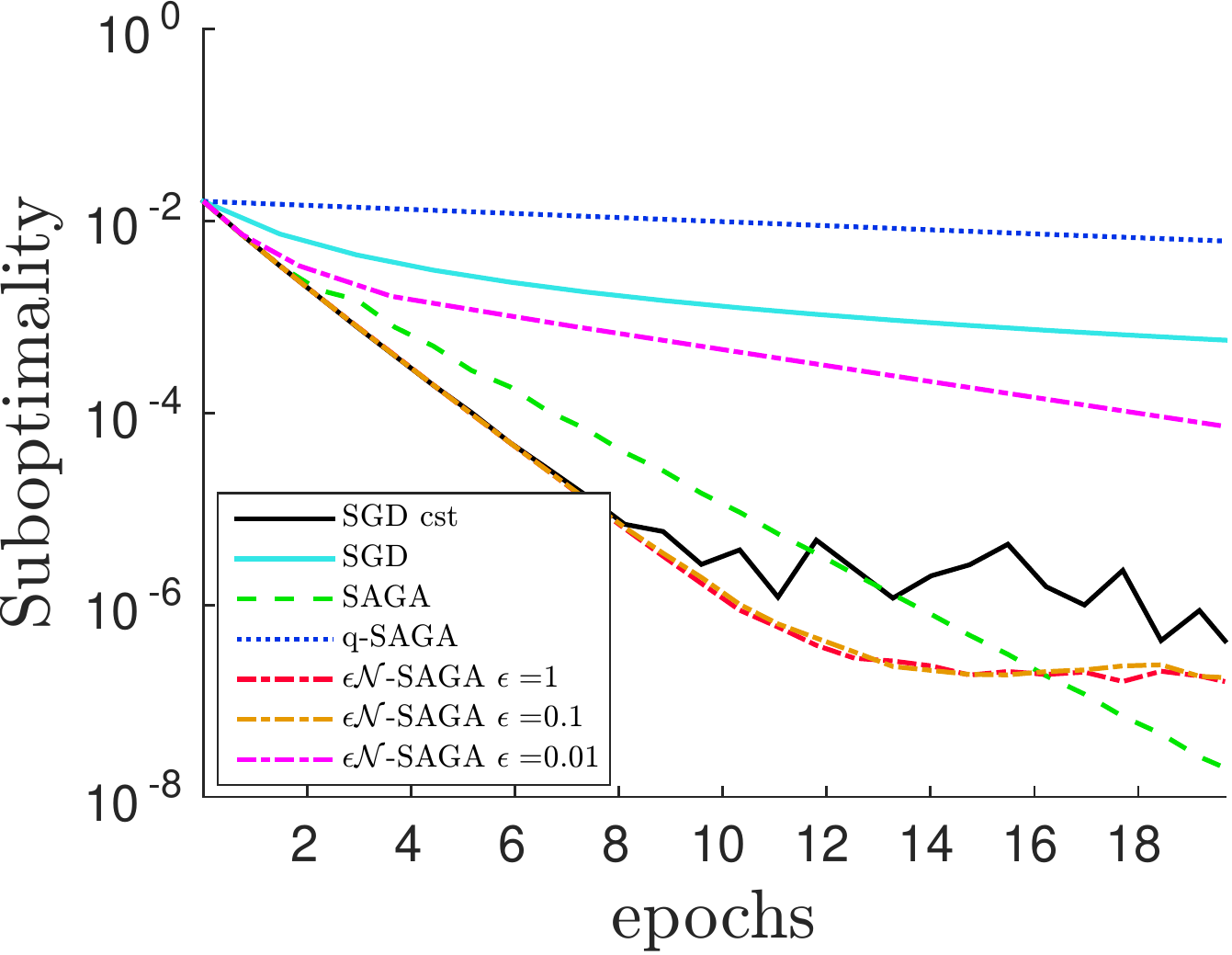} &
      \includegraphics[width=0.32\linewidth]{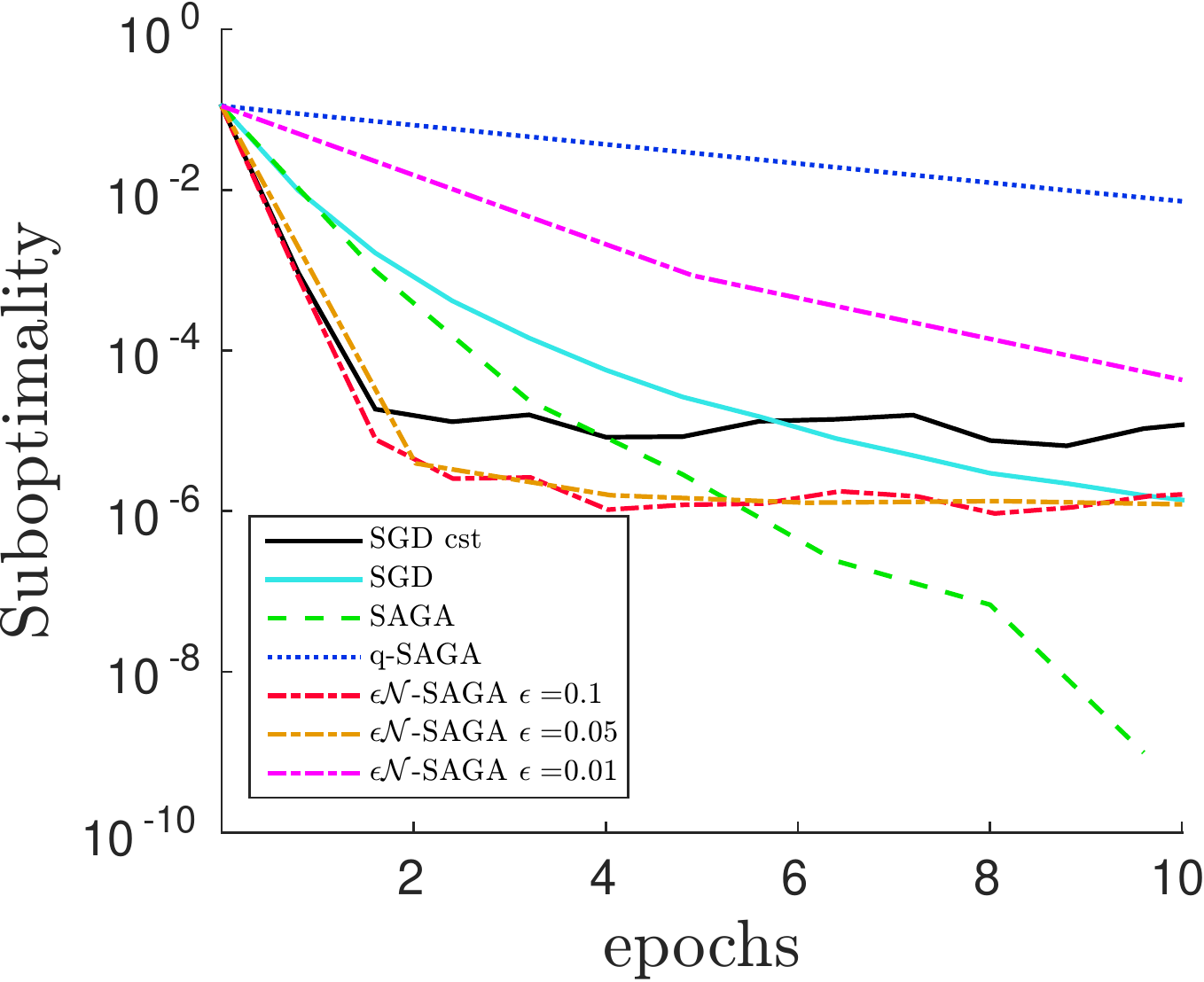} &
      \includegraphics[width=0.32\linewidth]{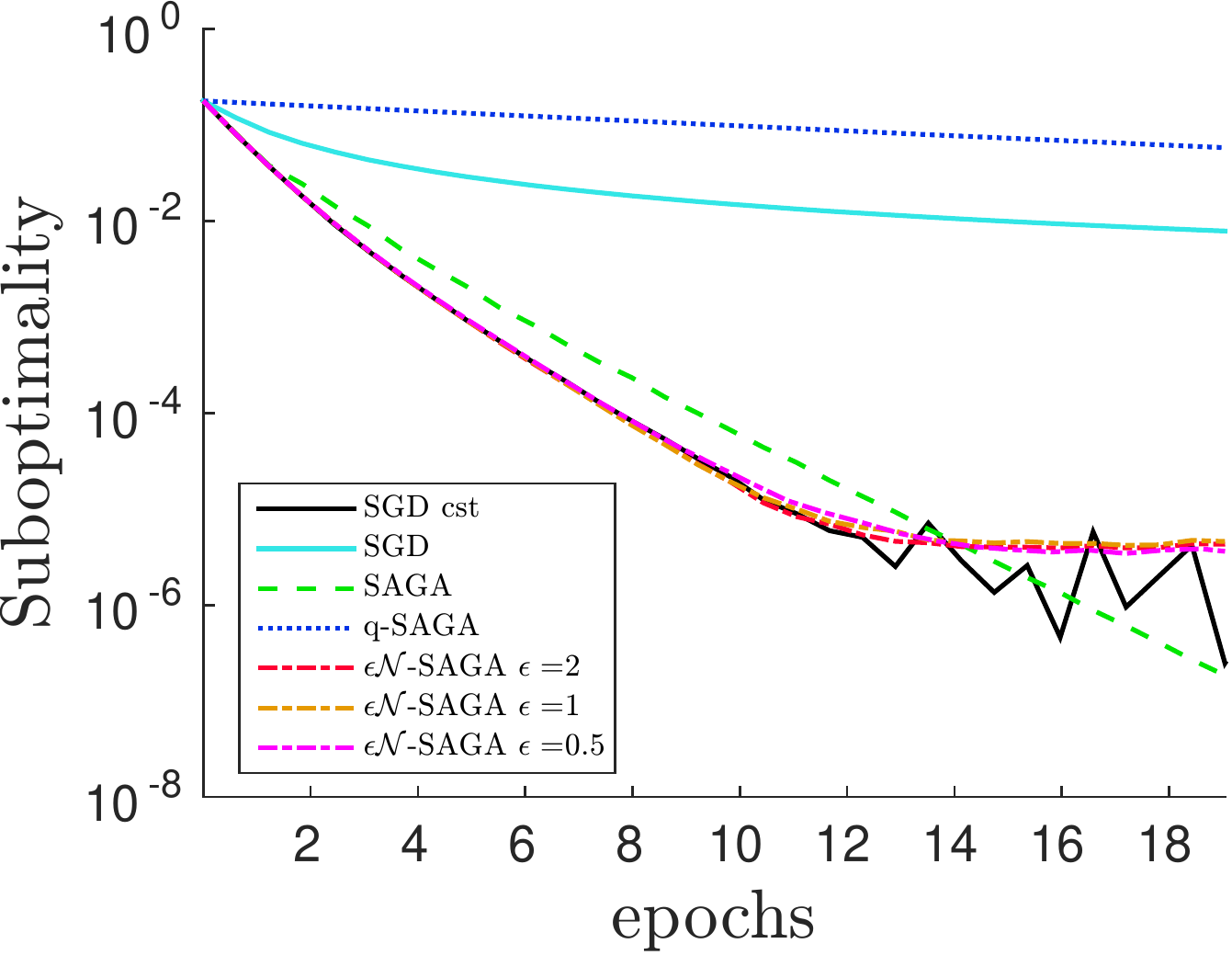} \\
      & {\scriptsize $\mu=10^{-1}$, gradient evaluation} & \vspace{2mm} \\ 
      \includegraphics[width=0.32\linewidth]{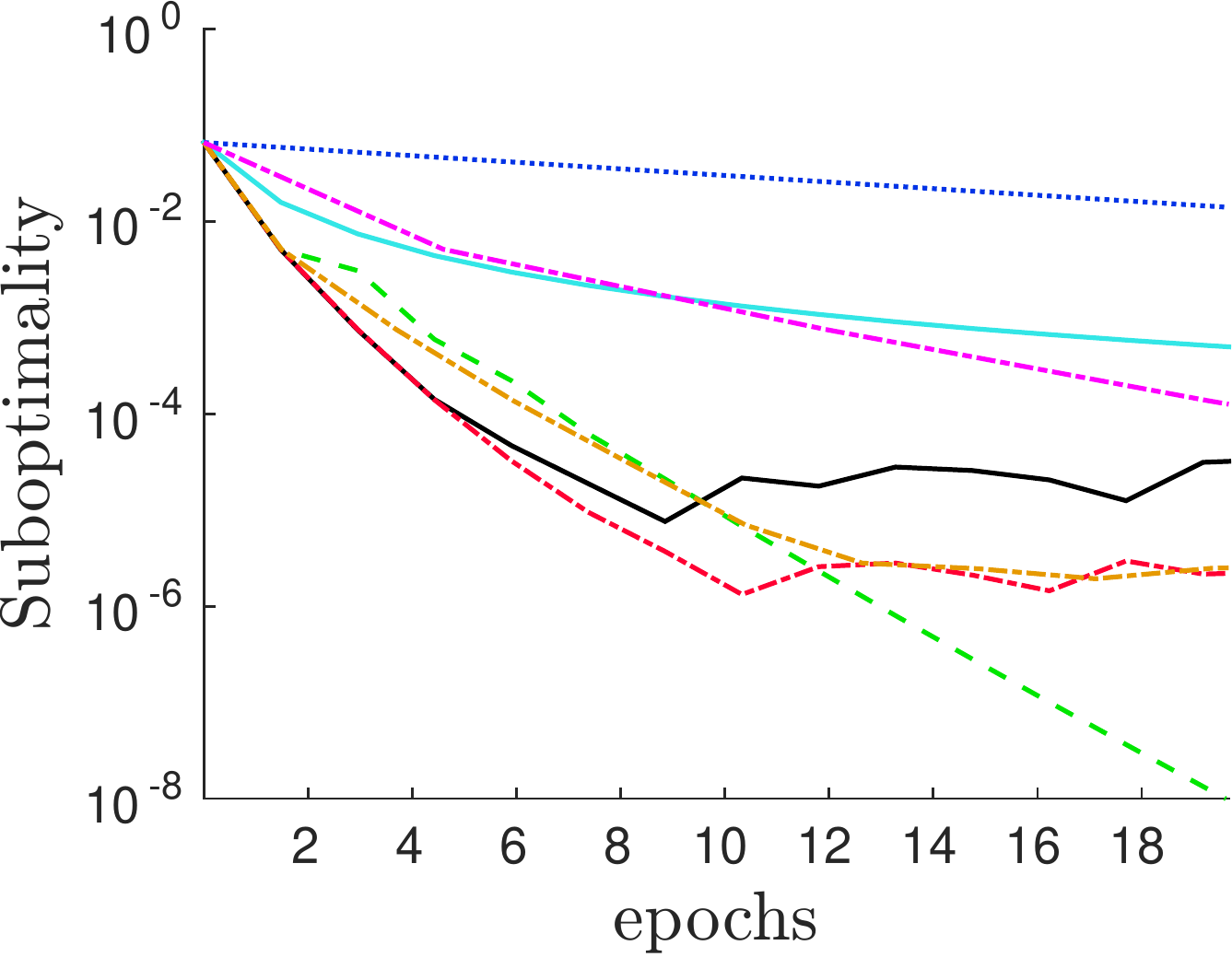} &
      \includegraphics[width=0.32\linewidth]{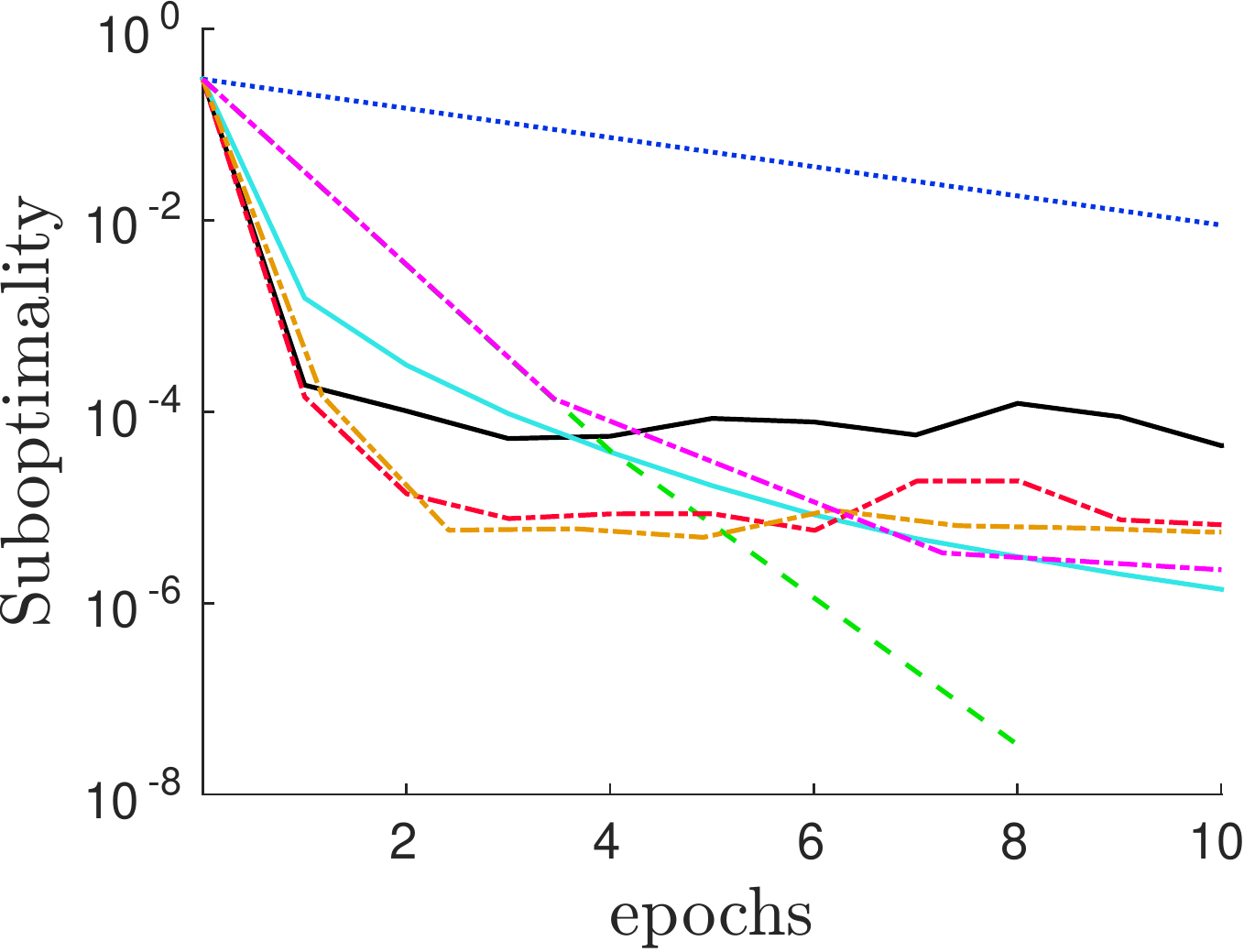} &
      \includegraphics[width=0.32\linewidth]{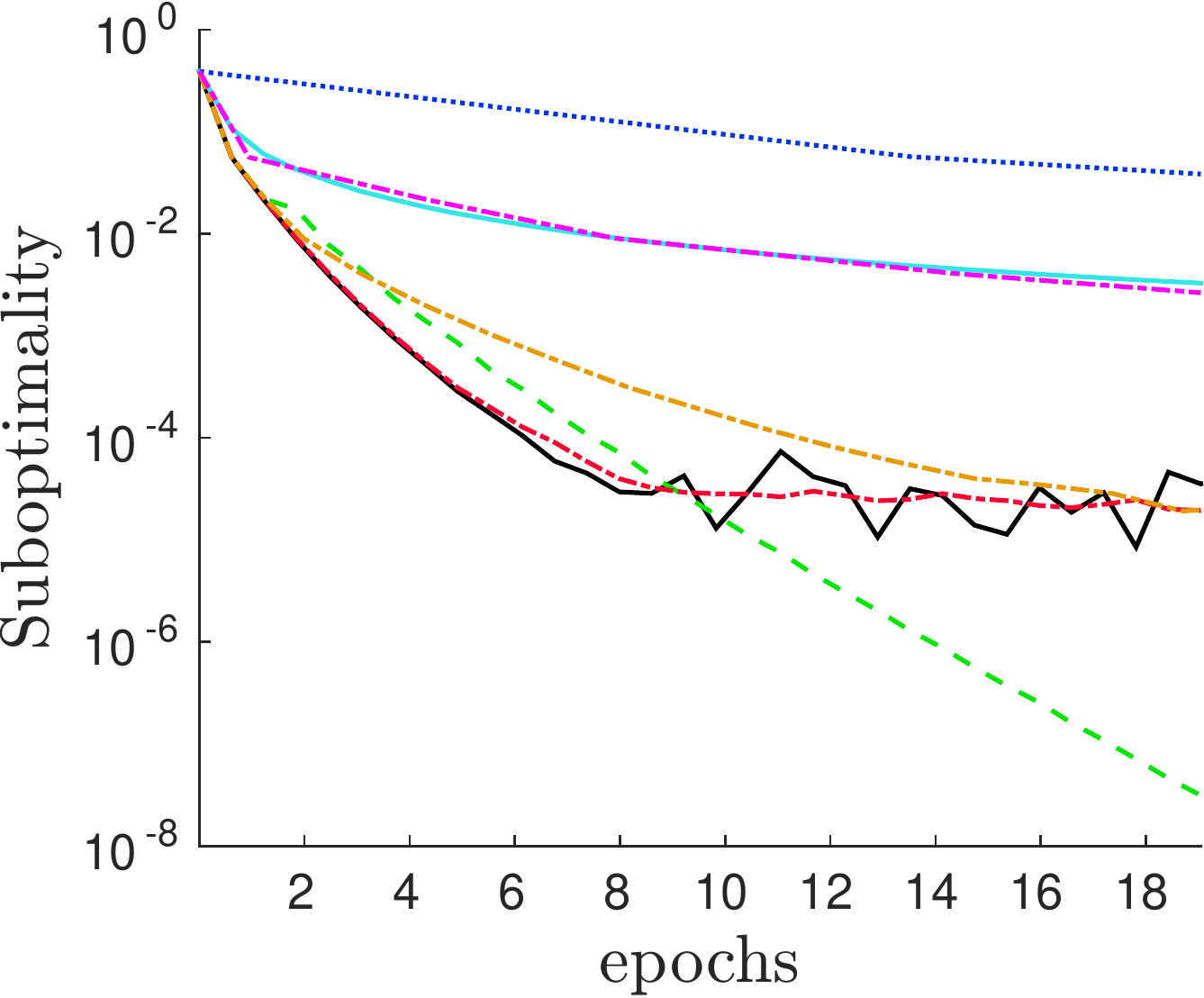} \\
      & {\scriptsize $\mu=10^{-3}$, gradient evaluation} & \vspace{2mm} \\
      \includegraphics[width=0.32\linewidth]{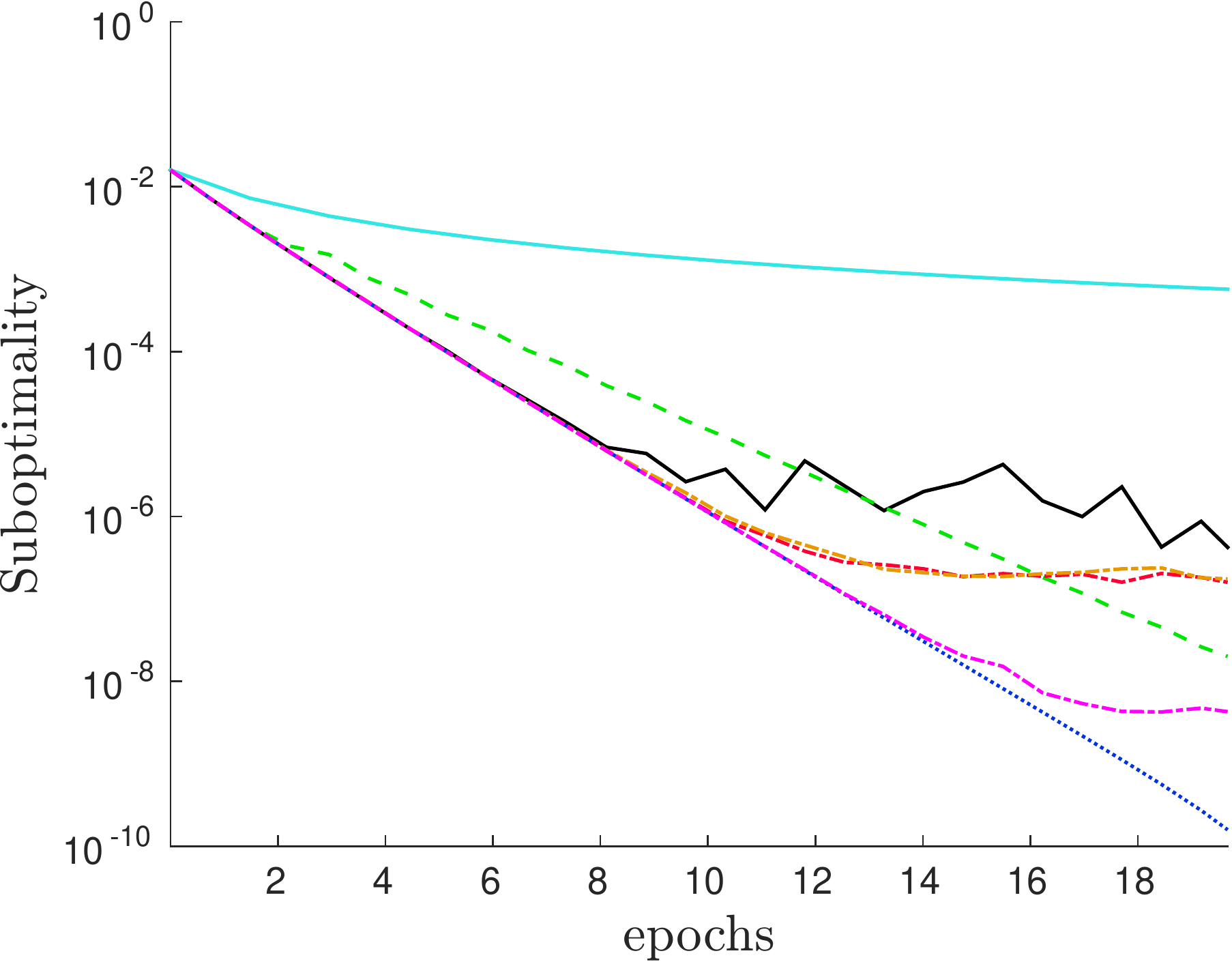} &
      \includegraphics[width=0.32\linewidth]{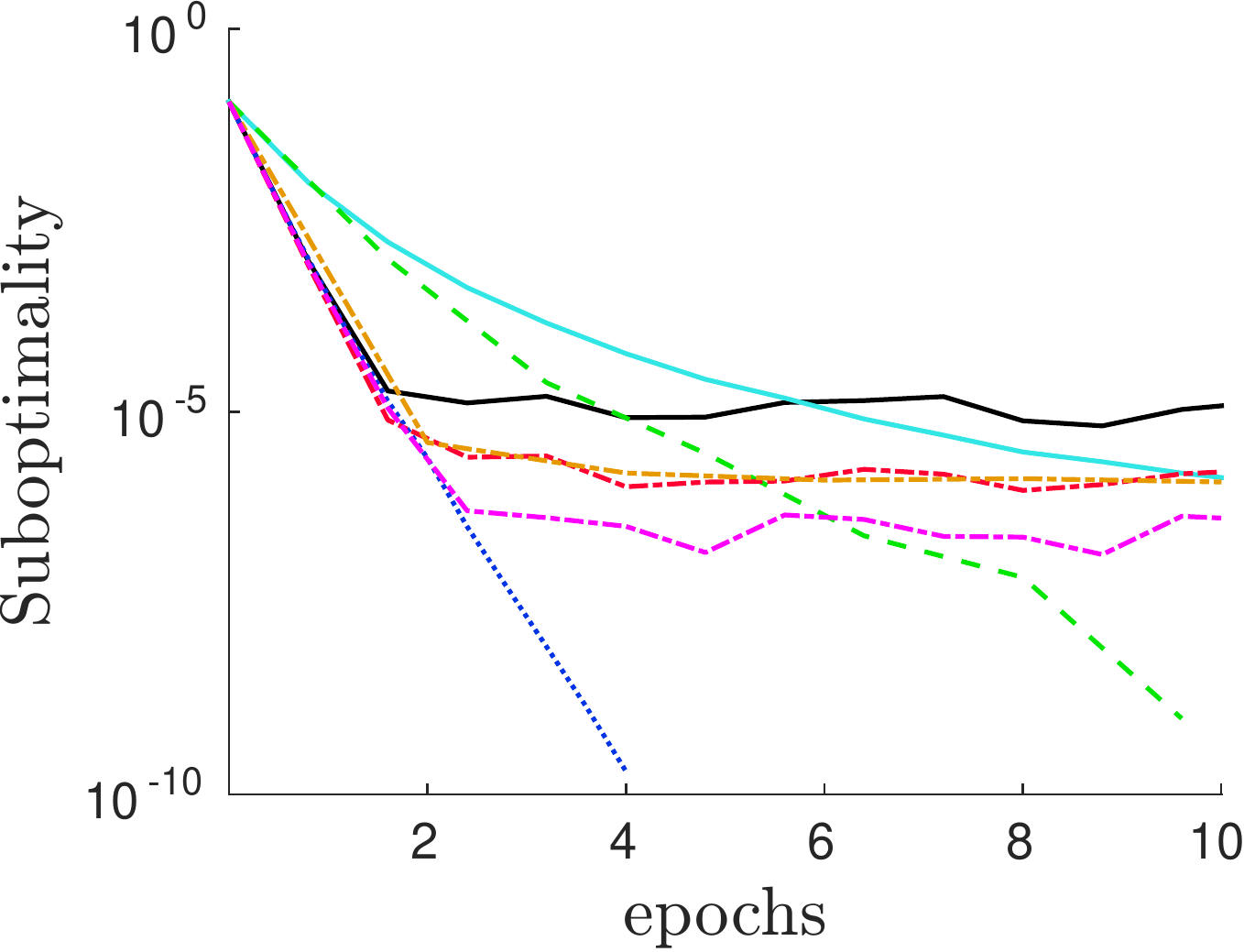} &
      \includegraphics[width=0.32\linewidth]{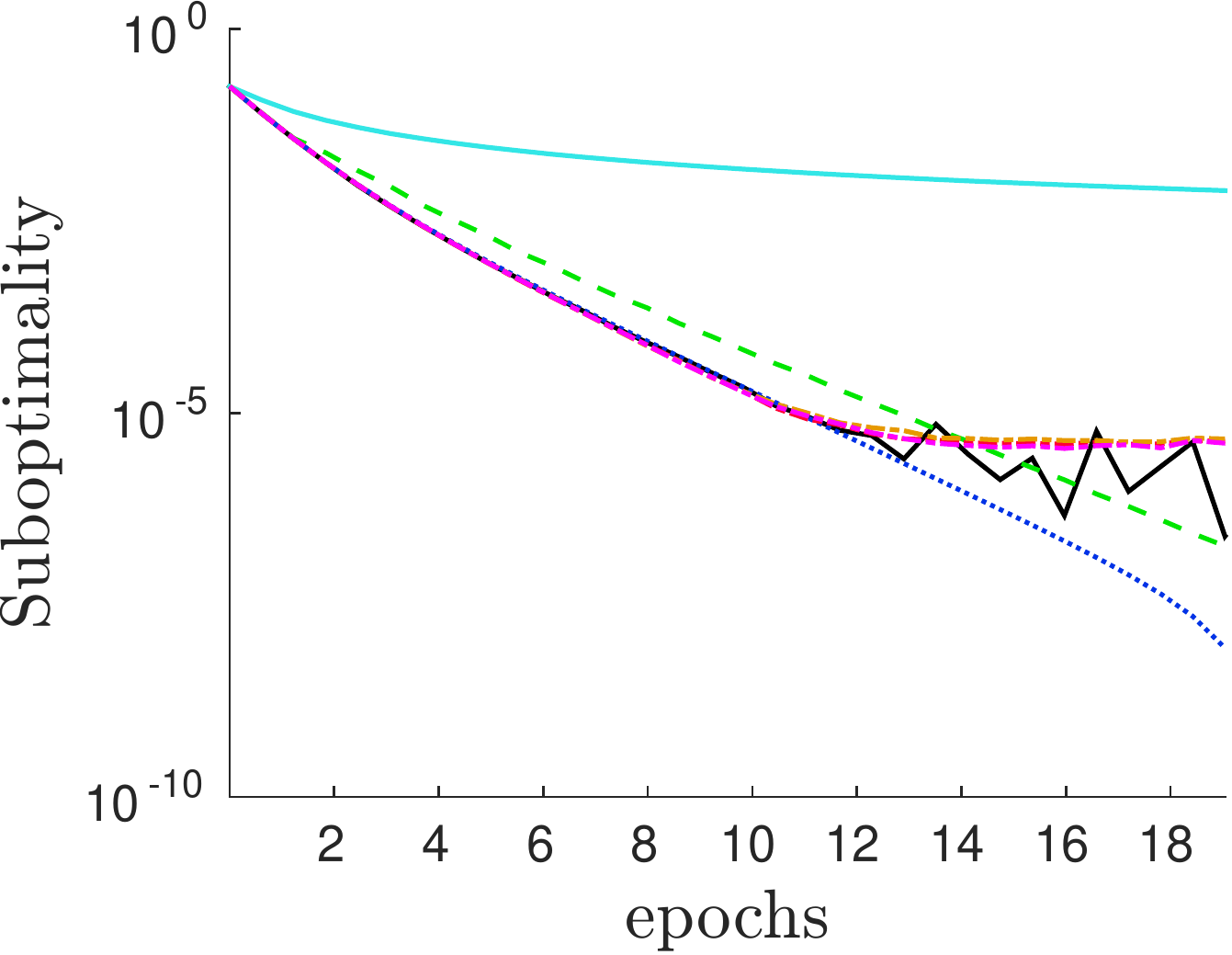} \\
      & {\scriptsize $\mu=10^{-1}$, datapoint evaluation} & \vspace{2mm} \\
      \includegraphics[width=0.32\linewidth]{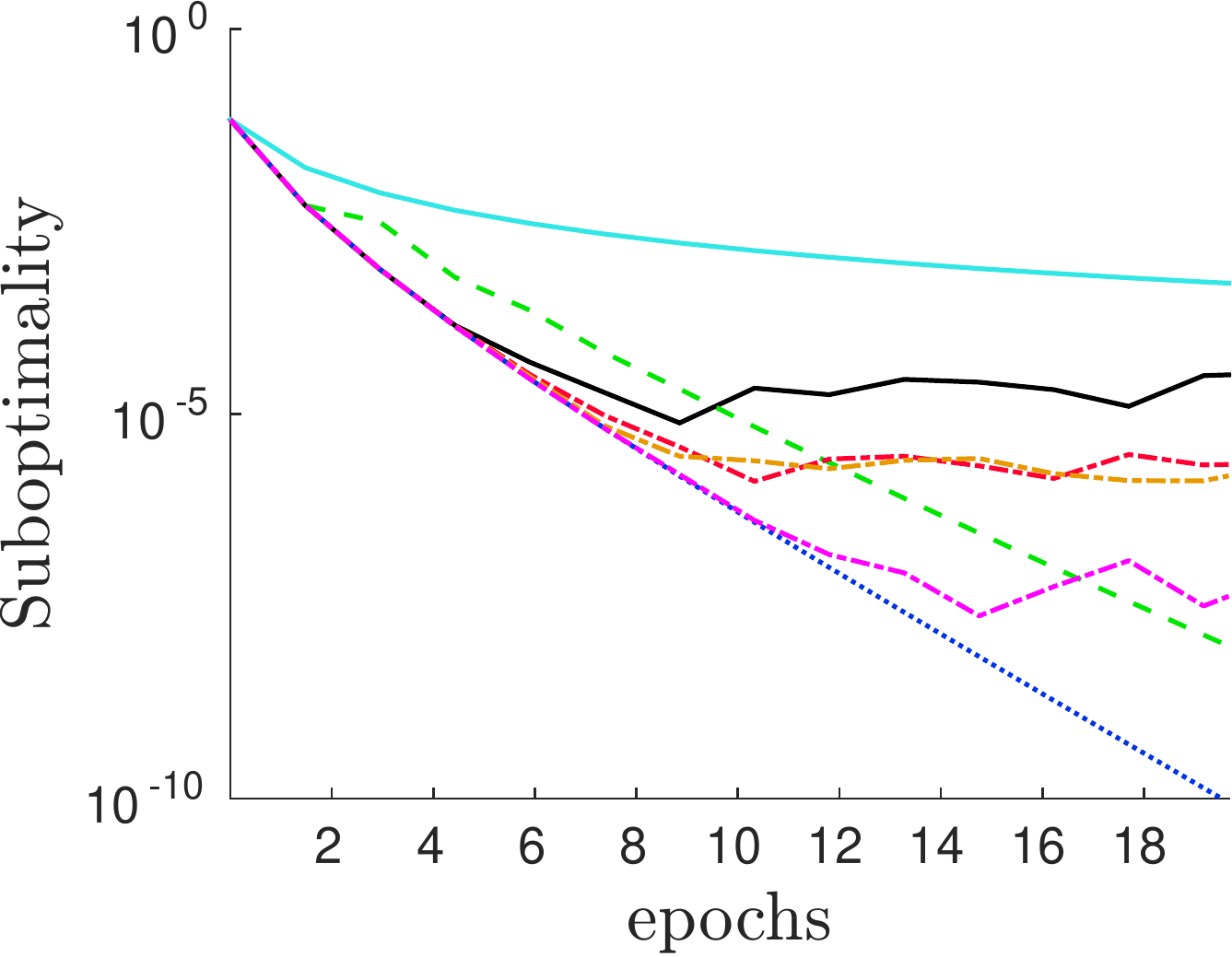} &
      \includegraphics[width=0.32\linewidth]{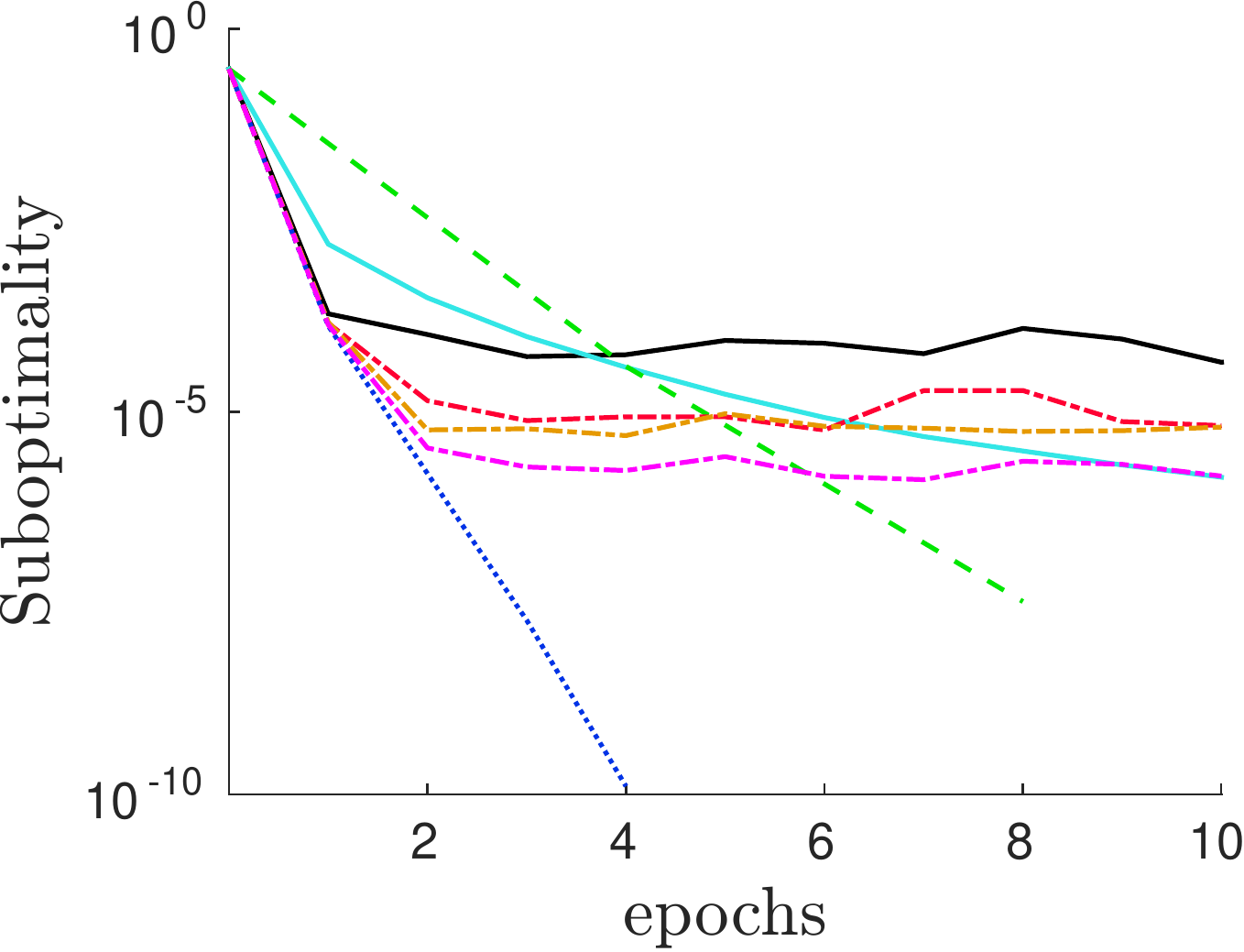} &
      \includegraphics[width=0.32\linewidth]{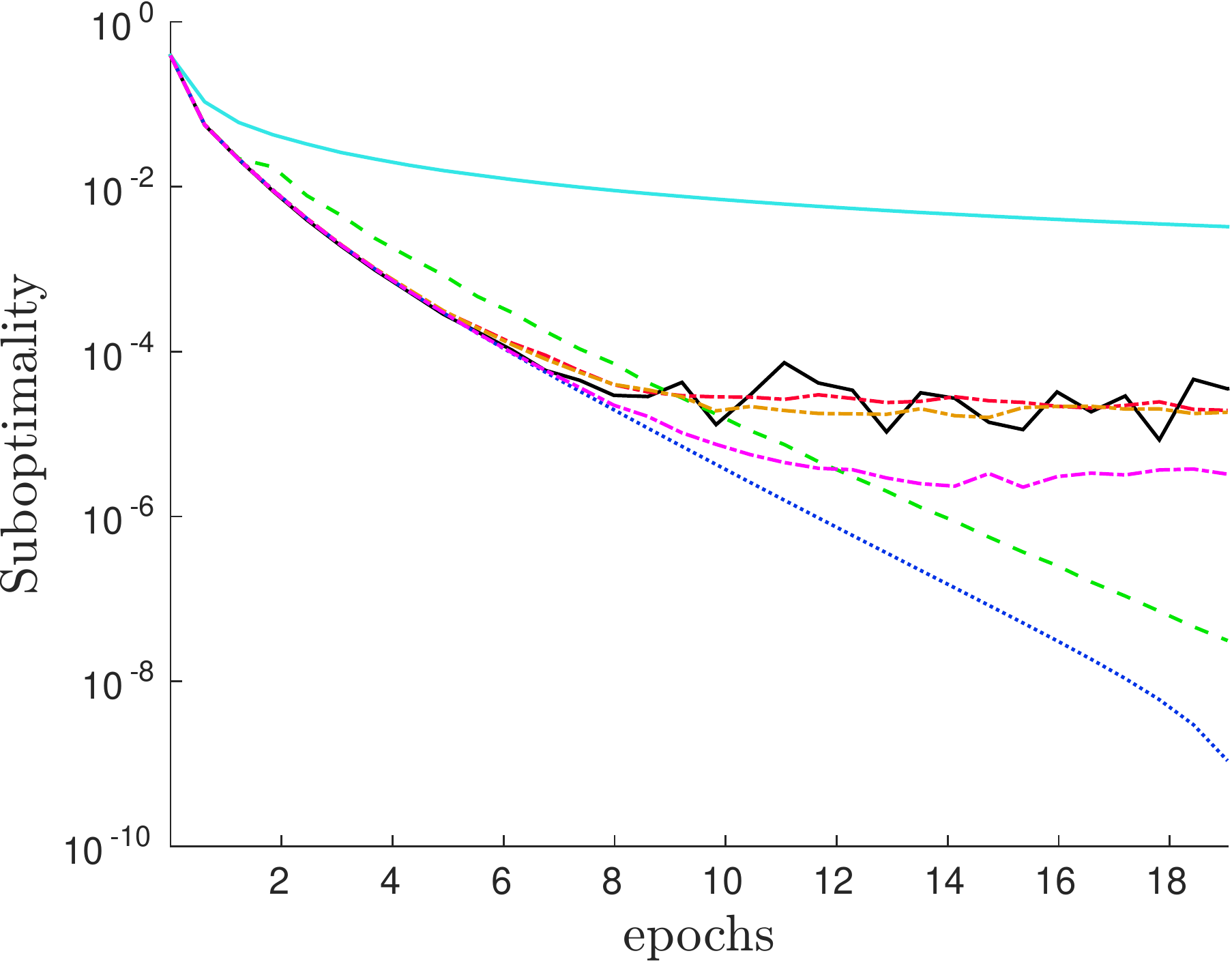} \\
      & {\scriptsize $\mu=10^{-3}$, datapoint evaluation} & \vspace{2mm} \\
    \end{tabular}
  \vspace{2mm}
  \caption{Comparison of $\epsilon {\cal N}$-SAGA, $q$-SAGA, SAGA and SGD (with decreasing and constant step size) on three datasets. The top two rows show the suboptimality as a function of the number of gradient evaluations for two different values of $\mu=10^{-1}, 10^{-3}$. The bottom two rows show the suboptimality as a function of the number of datapoint evaluations (i.e. number of stochastic updates) for two different values of $\mu=10^{-1}, 10^{-3}$.}
  \label{fig:results}
\end{center}
\end{figure*}

\section{Experimental Results}
\label{section:results}

\paragraph{Algorithms} We present experimental results on the performance of the different variants of memorization algorithms for variance reduced SGD as discussed in this paper. SAGA has been uniformly superior to SVRG in our experiments, so we compare SAGA and $\epsilon {\cal N}$-SAGA (from Eq.~\eqref{eq:en-saga}), alongside with SGD as a straw man and $q$-SAGA as a point of reference for speed-ups. We have chosen $q=20$ for $q$-SAGA and $\epsilon {\cal N}$-SAGA. The same setting was used across all data sets and experiments.

\paragraph{Data Sets}

As special cases for the choice of the loss function and regularizer in Eq.~\eqref{eq:problem}, we consider two commonly occurring problems in machine learning, namely least-square regression and $\ell_2$-regularized logistic regression.
We apply least-square regression on the million song year regression from the UCI repository. This dataset contains $n = 515,345$ data points, each described by $d=90$ input features.
We apply logistic regression on the {\it cov} and {\it ijcnn1} datasets obtained from the {\it libsvm} website
\footnote{\url{http://www.csie.ntu.edu.tw/~cjlin/libsvmtools/datasets}}. The
         {\it cov} dataset contains $n = 581,012$ data points, each
         described by $d = 54$ input features. The {\it ijcnn1}
         dataset contains $n = 49,990$ data points, each described by
         $d = 22$ input features.
We added an $\ell_2$-regularizer $\Omega(w) = \mu \|w\|_2^2$ to ensure
the objective is strongly convex.

\paragraph{Experimental Protocol}

We have run the algorithms in question in an i.i.d.~sampling setting and averaged the results over 5 runs. Figure \ref{fig:results} shows the  evolution of the suboptimality $f^\delta$ of the objective as a function of two different metrics: (1) in terms of the number of update steps performed (``datapoint evaluation"), and (2) in terms of the number of gradient computations (``gradient evaluation"). Note that SGD and SAGA compute one stochastic gradient per update step unlike $q$-SAGA, which is included here not as a practically relevant algorithm, but as an indication of potential improvements that could be achieved by fresher corrections. A step size $\gamma=\frac {q}{\mu n}$ was used everywhere, except for ``plain SGD''. Note that as $K \ll 1$ in all cases, this is close to the optimal value suggested by our analysis; moreover, using a step size of $\sim \frac{1}{L}$ for SAGA as suggested in previous work~\cite{schmidt2013minimizing} did not appear to give better results. For plain SGD, we used a schedule of the form $\gamma_t = \gamma_0/t$ with constants optimized coarsely via cross-validation. The $x$-axis is expressed in units of $n$ (suggestively called "epochs"). 

\vspace{-1.3mm}
\paragraph{SAGA vs.~SGD cst}  As we can see, if we run SGD with the same constant step size as SAGA, it takes several epochs until SAGA really shows a significant gain. The constant step-size variant of SGD is faster in the early stages until it converges to a neighborhood of the optimum, where individual runs start showing a very noisy behavior. 

\vspace{-1.3mm}
\paragraph{SAGA vs.~$q$-SAGA} 
$q$-SAGA outperforms plain SAGA quite consistently when counting stochastic update steps. This establishes optimistic reference curves of what we can expect to achieve with $\epsilon {\cal N}$-SAGA. The actual  speed-up is somewhat data set dependent.  

\vspace{-1.3mm}
\paragraph{$\epsilon{\cal N}$-SAGA vs.~SAGA and $q$-SAGA} $\epsilon {\cal N}$-SAGA with sufficiently small $\epsilon$ can realize much of the possible freshness gains of $q$-SAGA and performs very similar for a few (2-10) epochs, where it traces nicely between the SAGA and $q$-SAGA curves. We see solid speed-ups on all three datasets for both $\mu=0.1$ and $\mu=0.001$. 

\vspace{-1.3mm}
\paragraph{Asymptotics} It should be clearly stated that running $\epsilon{\cal N}$-SAGA at a fixed $\epsilon$ for longer will not result in good  asymptotics on the empirical risk. This is because, as theory predicts, $\epsilon{\cal N}$-SAGA can not drive the suboptimality to zero, but rather levels-off at a point determined by $\epsilon$. In our experiments, the cross-over point with SAGA was typically after $5-15$ epochs. Note that the gains in the first epochs can be significant, though. In practice, one will either define a desired accuracy level and choose $\epsilon$ accordingly or one will switch to SAGA for accurate convergence. 
\vspace{-1mm}

\section{Conclusion}

We have generalized variance reduced SGD methods under the name of memorization algorithms and presented a corresponding analysis, which commonly applies to all such methods. We have investigated in detail the range of safe step sizes with their corresponding geometric rates as guaranteed by our theory. This has delivered a number of new insights, for instance about the trade-offs between small ($\sim \frac 1n$) and large ($\sim \frac 1{4L})$ step sizes in different regimes as well as about the role of the freshness of stochastic gradients evaluated at past  iterates.

We have also investigated and quantified the effect of additional errors in the variance correction terms on the convergence behavior. Dependent on how $\mu$ scales with $n$, we have shown that such errors can be tolerated, yet, for small $\mu$, may have a negative effect on the convergence rate as much smaller step sizes are needed to still guarantee convergence to a small region. 
We believe this result to be relevant for a number of approximation techniques in the context of variance reduced SGD. 

Motivated by these insights and results of our analysis, we have proposed ${\epsilon \cal N}$-SAGA, a modification of SAGA that exploits similarities between training data points by defining a neighborhood system. Approximate versions of  per-data point gradients are then computed by sharing information among neighbors. This opens-up the possibility of variance-reduction in a streaming data setting, where each data point is only seen once. We believe this to be a promising direction for future work. 

Empirically, we have been able to achieve consistent speed-ups for the initial phase of regularized risk minimization. This shows that approximate computations of variance correction terms constitutes a promising approach of trading-off computation with solution accuracy. 

\paragraph{Acknowledgments} We would like to thank Yannic Kilcher, Martin Jaggi, R\'{e}mi Leblond and the anonymous reviewers for helpful suggestions and corrections.

\bibliographystyle{abbrv}
\bibliography{nsaga}

\begin{thebibliography}{10}

\bibitem{andoni2018LSH}
A.~Andoni and P.~Indyk.
\newblock Near-optimal hashing algorithms for approximate nearest neighbor in
  high dimensions.
\newblock {\em Commun. ACM}, 51(1):117--122, 2008.

\bibitem{bottou2010}
L.~Bottou.
\newblock Large-scale machine learning with stochastic gradient descent.
\newblock In {\em COMPSTAT}, pages 177--186. Springer, 2010.

\bibitem{dasgupat2015RPT}
S.~Dasgupta and K.~Sinha.
\newblock Randomized partition trees for nearest neighbor search.
\newblock {\em Algorithmica}, 72(1):237--263, 2015.

\bibitem{defazio2014}
A.~Defazio, F.~Bach, and S.~Lacoste-Julien.
\newblock {SAGA}: A fast incremental gradient method with support for
  non-strongly convex composite objectives.
\newblock In {\em Advances in Neural Information Processing Systems}, pages
  1646--1654, 2014.

\bibitem{johnson2013}
R.~Johnson and T.~Zhang.
\newblock Accelerating stochastic gradient descent using predictive variance
  reduction.
\newblock In {\em Advances in Neural Information Processing Systems}, pages
  315--323, 2013.

\bibitem{konevcny2013}
J.~Kone{\v{c}}n{\`y} and P.~Richt{\'a}rik.
\newblock Semi-stochastic gradient descent methods.
\newblock {\em arXiv preprint arXiv:1312.1666}, 2013.

\bibitem{robbins1951}
H.~Robbins and S.~Monro.
\newblock A stochastic approximation method.
\newblock {\em The annals of mathematical statistics}, pages 400--407, 1951.

\bibitem{schmidt2014convergence}
M.~Schmidt.
\newblock Convergence rate of stochastic gradient with constant step size.
\newblock {\em UBC Technical Report}, 2014.

\bibitem{schmidt2013minimizing}
M.~Schmidt, N.~L. Roux, and F.~Bach.
\newblock Minimizing finite sums with the stochastic average gradient.
\newblock {\em arXiv preprint arXiv:1309.2388}, 2013.

\bibitem{shalev2011}
S.~Shalev-Shwartz, Y.~Singer, N.~Srebro, and A.~Cotter.
\newblock Pegasos: Primal estimated sub-gradient solver for {SVM}.
\newblock {\em Mathematical programming}, 127(1):3--30, 2011.

\bibitem{shalev2013stochastic}
S.~Shalev-Shwartz and T.~Zhang.
\newblock Stochastic dual coordinate ascent methods for regularized loss.
\newblock {\em The Journal of Machine Learning Research}, 14(1):567--599, 2013.

\end{thebibliography}

\newpage

\appendix

\section{Appendix}
\label{sec:appendix}

\setcounter{lemma}{0}
\setcounter{theorem}{0}
\setcounter{corollary}{0}

\begin{lemma}
For the iterate sequence of any algorithm that evolves solutions according to Eq.~\eqref{eq:sgd-corrected}, the following holds for a single update step, in expectation over the choice of $i$, with \\$\triangle := \| w - w^*\|^2  - \E \| w^+ - w^*\|^2$, then: 
\begin{align}
\triangle \ge  &\quad \gamma \mu \| w- w^*\|^2  - 2 \gamma^2  \E \| \alpha_i - f_i'(w^*)\|^2  + \left( 2\gamma - 4 \gamma^2 L  \right) f^\delta(w) \,.
\nonumber %
\end{align}
\vspace*{-6mm}
\begin{proof} 
Starting from Eq.~\eqref{eq:recurrence} we have
\begin{align*}
\triangle
= &
2 \gamma \langle f'(w), w-w^*\rangle - \gamma^2 \E \| g_i(w) \|^2 
\\ \stackrel{\eqref{eq:naive-strong-convexity}}{\ge} &  
\gamma \mu \| w-w^*\|^2 + 2 \gamma f^\delta (w)  - \gamma^2 \E \| g_i(w) \|^2 \\
\stackrel{\eqref{eq:beta-split}}{\ge} &  
\gamma \mu  \| w-w^*\|^2  +  2 \gamma f^\delta (w) 
- 2 \gamma^2 \E \| f'_i(w) - f'_i(w^*) \|^2 
-2 \gamma^2  \E \| \bar \alpha_i - f_i'(w^* )\|^2 \\
\stackrel{\eqref{eq:expected-smoothness},\eqref{eq:alpha-simplification}}{\ge} &  
\gamma \mu \| w- w^*\|^2  + 
2 \gamma ( 1 - 2\gamma L  )  f^\delta(w)
  - 2 \gamma^2 \E \| \alpha_i - f_i'(w^*)\|^2  \nonumber
\,.\end{align*}
\end{proof}
\end{lemma}

\begin{lemma} 
For a uniform $q$-memorization algorithm, it holds that 
\begin{align*}
\E \bar H^+ = \left( \frac{n-q}{n} \right) \bar H + \frac{2Lq}{n} \, f^\delta(w).
\end{align*}
\begin{proof} From the uniformity property $(*)$ in Definition \ref{def:memorization}, it follows that 
\begin{align}
n \E H^+ \! = \sum_{i=1}^n \E H^+_i 
\stackrel{(*)}= \! \sum_{i=1}^n \left( \left(1-\frac{q}{n} \right) H_i +  2L\left( \frac{q}{n} \right)h_i(w) \right)
= (n-q) \bar H + \frac {2L q}n \sum_{i=1}^n h_i(w).
\nonumber
\end{align}
Exploiting the fact that $\frac 1n \sum_{i=1}^n h_i(w) = f(w)-f(w^*) + 0= f^\delta(w)$  completes the proof. 
\end{proof} 
\label{lemma:hrecurrence2-appendix}
\end{lemma}

\begin{lemma}
\label{lemma:lyapunov-appendix}
Fix $c \in (0;1]$ and $\sigma \in [0;1]$ arbitrarily. For any uniform $q$-memorization algorithm with sufficiently small step size $\gamma$ such that
\begin{align*}
\gamma \le \frac{1}{L} \min \left\{\frac{K\sigma }{2K + 4c\sigma}, \frac{1-\sigma}{2} \right\}, \quad 
\text{and} \quad K:=  \frac{4qL}{n \mu},
\end{align*}
we have that
\begin{align}
\E \L_\sigma(w^+,H^+)  \le (1-\rho) \L_\sigma(w,H),\quad \text{with} \quad \rho := c \mu \gamma.
\end{align}
Note that $\gamma < \frac 1{2L} \max_{\sigma \in [0,1]} \min\{ \sigma ,  1-\sigma\} = \frac{1}{4L}$ (in the $c \to 0$ limit).
\begin{proof}
From Lemma \ref{lemma:w-recurrence}, we can see that we will have $\rho \leq \gamma \mu$ based on the $ \| w - w^*\|^2$ part of $\L_\sigma$. Hence, we can write the rate as $\rho = c \mu \gamma$, where  $0 < c \leq 1$. 

Let us now apply both, Lemma \ref{lemma:w-recurrence} and Lemma \ref{lemma:hrecurrence2}, to quantify the progress guaranteed to be made in one iteration of the algorithm in expectation, combining the changes to the iterate $w \to w^+$ as well as those to the memory $\alpha \to \alpha^+$ into $\L_\sigma$. Set $\triangle_\sigma := \L_\sigma(w,H) - \E \L_\sigma^+(w,H)$, then 
\begin{align}
\triangle_\sigma  = &  \| w- w^*\|^2 - \E \| w^+ - w^*\|^2 + S \sigma \,  \left( \bar H - \E \bar H^+ \right) \\
\ge &  \gamma \mu \| w- w^*\|^2  - 2 \gamma^2  \E \| \alpha_i - f_i'(w^*)\|^2  + 2\gamma \left( 1 - 2\gamma L  \right) f^\delta(w) \nonumber \\
& + S \sigma \left( \bar H -  \left( \frac{n-q}{n} \right) \bar H - \frac{2Lq}{n} \, f^\delta(w) ) \right).
\nonumber
\end{align}
As we argued after Eq.~\eqref{eq:recurrence-h1}, the definition of $H_i$ combined with property~\eqref{eq:alpha-bound} ensure the crucial bound $\E\| \alpha_i - f_i'(w^*) \|^2 \le \bar H$. Including it and gathering terms in the same ``units", we get: 
\begin{align}
\triangle_\sigma \ge &  \; \gamma \mu \| w- w^*\|^2 + 
\left[ S \sigma \left( \frac qn \right) - 2 \gamma^2  \right] \bar H 
 \label{eq:recbrackets2}
+ 2 \left[ 
   -S \sigma  L \left( \frac qn \right) 
  + \gamma \left( 1 -2 \gamma L \right) 
\right] f^\delta(w) 
\end{align}
We can further simplify the term in the second rectangular brackets with the definition of $S$ (in hindsight motivating its definition):
\begin{align}
- 2S \sigma  L \left( \frac qn \right)  + 2\gamma \left( 1 - 2\gamma L\right)
= 2\gamma \left[ -\sigma + \left( 1 - 2 \gamma L \right) \right] 
\end{align}
We require this term to be non-negative, so that we can safely drop it. This leads an upper bound requirement on the step size:
\begin{align}
- \sigma + 1 - 2\gamma L \ge 0 \iff 
\gamma \leq \frac{1- \sigma}{2 L}.
\end{align}
The term in the first rectangular brackets in Eq.~\eqref{eq:recbrackets2} needs to be $\ge \rho S \sigma$ in order to recover $\rho \L_\sigma = \rho \left( \| w- w^*\|^2 + S \sigma \bar H \right)$. Inserting the definition of $S$, $\rho$ and dividing by $\gamma$ yields 
\begin{align}
\frac {\sigma}{L} - 2 \gamma \ge \frac{\rho S \sigma}{\gamma}
 = c \gamma \mu \frac{n \sigma}{L q} = \frac{4c  \sigma \gamma}{K}
& \iff \gamma \le  \frac 1L \frac{K \sigma}{2 K + 4 c \sigma}
\end{align}
We can summarize the derivation in the claimed combined inequality.
\end{proof} 
\end{lemma}

\begin{theorem}
\label{theorem:main-appendix}
Consider  a uniform $q$-memorization algorithm. For any step size $\gamma = \frac a {4L}$, with $a<1$ the algorithm converges at a geometric rate of at least $(1-\rho(\gamma))$ with 
\begin{align*}
\rho(\gamma) = \frac{q}{n} \cdot \frac{1 - a}{1-a/2} 
= \frac{\mu}{4L} \cdot \frac{K (1 - a)}{1-a/2}, \;\;  \text{if} \;\gamma \geq \gamma^*(K), \;\; 
\text{otherwise} \;\; \rho(\gamma) = \mu \gamma   \;\; 
\end{align*}
where 
\begin{align*}
\gamma^*(K) :=  \frac {a^*(K)} {4 L}, \quad 
a^*(K) := \frac{2K}{1+K + \sqrt{1+K^2}},
\quad K:=  \frac{4qL}{n \mu}
\end{align*}
\begin{proof} 
Consider a fixed $\gamma < \frac 1{4L}$. There are potentially (infinitely) many choices of $(c,\sigma)$ that fulfill the condition in Eq.~\eqref{eq:gamma-admissible}. Among those, the largest rate is obtained by maximizing $c \le 1$ as $\rho(\gamma) = c \mu \gamma$. Note that for any $\gamma$ that does not achieve Eq.~\eqref{eq:gamma-admissible} with equality for both terms, one can find a larger $\gamma$ with the same choice of $c$ by either increasing (slack in the first inequality) or decreasing (slack in the second inequality) $\sigma$. We thus focus on step sizes that are maximal for some choice of $(c,\sigma)$. Equality with the second bound directly gives us
\begin{align} \label{eq:sigma-star}
\frac 1 L \frac{1-\sigma}{2} \stackrel != \gamma \; \Longrightarrow \; \sigma^*  = 1 - 2L \gamma.
\end{align} We plug this into the first bound and again equal $\gamma$, which yields an optimality condition for $c$
\begin{align}
& L \gamma  \stackrel != \frac{K \sigma^*}{2K + 4c \sigma^* } 
 \iff   c^*  %
= \frac K{4\gamma L} \left[ 1 - \frac{2 \gamma L}{\sigma^*} \right]
\Longrightarrow c^* = \frac K{4\gamma L}  \frac{1 - 4 L \gamma}{1 - 2L \gamma}
\label{eq:c-optimal}
\end{align}
and thus
\begin{align}
& \rho = c \mu \gamma = \frac{\mu K}{4L} \frac{1 - 4 L \gamma}{1 - 2L \gamma}
= \frac{q}{n}  \frac{1 - 4 L \gamma}{1 - 2L \gamma}
\end{align}
It remains to check what the admissible range of $\gamma$ is that achieves the bound in Eq.~\eqref{eq:gamma-admissible} as we required. The latter is determined by the constraints $c \in (0;1]$.  From Eq.~\eqref{eq:c-optimal} we can read off for $c^*>0$, 
\begin{align}
1 - 4 L \gamma > 0  \iff \gamma < \frac {1}{4L} \,.
\end{align}
At the other extreme of $c=1$ we can solve the resulting quadratic equation in $\gamma$ 
\begin{align}
\gamma =  \frac{q}{n \mu}  \frac{1 - 4 L \gamma}{1 - 2L \gamma} = 
\frac{K}{4L} \frac{1 - 4 L \gamma}{1 - 2L \gamma} 
\end{align}
to get $\gamma= \gamma^*(K)$ as claimed in Eq.~\eqref{eq:gammalk} (excluding the second root which yields $\gamma > \frac{1}{4L}$). Moreover, for $\gamma < \gamma^*(K)$  we choose $c=1$ to maximize the rate and have $\rho = \mu \gamma$.
\end{proof}
\end{theorem}

\begin{corollary}
\label{corollary:opt-rho-appendix}
In Theorem \ref{theorem:main}, $\rho$ is maximized for $\gamma = \gamma^*(K)$. We can write $\rho^*(K) = \rho(\gamma^*)$ as 
\begin{align*}
\rho^*(K) = 
\frac \mu {4L} a^*(K) %
= \frac{q}{n}\frac{a^*(K)}{K} = \frac qn \left[ \frac{2}{1+K + \sqrt{1 + K^2} } \right]
\end{align*}
In the big data regime $\rho^* = \frac qn (1-\frac 12 K + O(K^3))$, whereas in the ill-conditioned case $\rho^* = \frac \mu{4L} (1 - \frac {1}{2}K^{-1} + O(K^{-3}))$.
\begin{proof}
Plugging in the definitions of $\gamma^*(K)$ and $K$ and performing some symbolic simplifications yields the result. 
\end{proof}
\end{corollary}

\begin{corollary}
\label{corollary:universal-appendix}
Choosing $\gamma = \frac{2 -\sqrt{2} }{4L}$, leads to $\rho(\gamma) \ge (2 - \sqrt{2}) \rho^* > \frac 12 \rho^*$.
\begin{proof} 
Write $\gamma = \frac{a}{4L}$, then if $\gamma \geq \gamma^*(K)$: $\frac{\rho}{\rho^*} \geq \frac{1 - a}{1- a/2}$, otherwise: $ \frac{\rho}{\rho^*} = \frac{\gamma}{\gamma^*} \ge  a$, with equality when $K = \infty$. Setting both equal yields $a =  2 - \sqrt{2} \approx 0.5858$. 
\end{proof} 
\end{corollary}

\begin{corollary}
Choosing $\gamma = \frac{a}{4L}$ with $ a<1$ yields $\rho = \min\{ \frac {1-a}{1 - \frac 12 a} \frac qn, \frac{a}{4} \frac \mu L \}$. In particular, we have for the choice $\gamma = \frac{1}{5L}$ that $\rho = \min\{ \frac {1}{3} \frac qn, \frac 15 \frac \mu L \}$.
\begin{proof} 
If $\gamma \ge \gamma^*(R)$ then $\rho = \frac{1-a}{1 -a/2}\frac qn = \frac{1}{3} \frac{q}{n}$; otherwise, $\rho = \mu \gamma = \mu \frac{a}{4L} = \frac{1}{5} \frac{\mu}{L}$.
\end{proof} 
\end{corollary}

\begin{theorem}
\label{theorem:approximate-appendix}
Consider a uniform $q$-memorization algorithm with $\alpha$-updates that are on average $\epsilon$-accurate (i.e. $\E\| \alpha_i - \beta_i \|^2 \le \epsilon$). For any step size $\gamma \le \tilde{\gamma}(K)$, where $\tilde{\gamma}$ is given in Eq.~\eqref{eq:atilde} in Corollary~\ref{corollary:patch} below (note that $\tilde{\gamma}(K) \geq \frac{2}{3} \gamma^*(K)$ and $\tilde{\gamma}(K) \to \gamma^*(K)$ as $K \to 0$), we get 
\begin{align}
\Efull \L(w^t,H^t) \le  (1-\mu \gamma)^t  \L_0+ \frac{4 \gamma \epsilon}{\mu}, \quad \text{ with } \L_0 := \|w^0 - w^*\|^2 + s(\gamma) \E \|f_i(w^*)\|^2 ,
\end{align}
where $\Efull$ denote the (unconditional) expectation over histories (in contrast to $\E$ which is conditional), and $ s(\gamma) := \frac{4 \gamma}{K \mu} (1-2 L \gamma)$.
\begin{proof} Following the same line of argument as in Lemma \ref{lemma:lyapunov} and Theorem \ref{theorem:main} with the modifications summarized in Corollary \ref{corollary:patch}
\begin{align*}
\E \L(w^+,H^+) \le  (1-\gamma\mu)  \L(w,H) + 4 \gamma^2 \epsilon
\end{align*}
and unrolling the recurrence over $t$ \vspace*{-5mm}
\begin{align*} 
    \Efull \L(w^t,H^t) \le  (1-\gamma\mu)^t  \L(w^0,H^0) + \overbrace{\left[ \sum_{s=0}^{t-1} (1- \gamma\mu)^s\right]}^{\leq 1/(\gamma \mu)} 4 \gamma^2 \epsilon
\end{align*}
using $\frac{1}{1-x} = \sum_{s=0}^\infty x^s$ applied with $x=(1-\rho)$ (see \cite{schmidt2014convergence} for its use for constant step size SGD). According to Eq.~\eqref{eq:lyapunov_fct}, $\L(w^0, H^0) = \|w-w^*\|^2 + S \sigma \bar{H}^0$, where $S = \frac{\gamma n}{L q} = \frac{4 \gamma}{K \mu}$. As the algorithm initializes $\alpha_i^0$ to $0$, we have $\bar{H}^0 = \E \|f'_i(w^*)\|^2$. Finally, the proof of Corollary~\ref{corollary:patch} follows the proof of Theorem~\ref{theorem:main-appendix} and also gives $\sigma^* = 1-2 L \gamma$ as in Eq.~\eqref{eq:sigma-star}. Substituting in $\L(w^0, H^0)$ gives $\L_0$.
\end{proof}
\end{theorem} 

\begin{corollary}
With $\gamma = \min\{\mu , \tilde{\gamma}(K)\}$ we have 
\begin{align*}
\frac {4 \gamma \epsilon }{\mu}  \le 4 \epsilon, \qquad 
\text{with a rate} \quad \rho = \min\{\mu^2, \mu \tilde{\gamma}\} \, .
\end{align*}
\begin{proof}
By definition, $\gamma \leq \mu$, thus $\gamma/ \mu \leq1$, yielding the first claim. By definition, we also have $\gamma \leq \tilde{\gamma}(K)$, thus from Theorem~\ref{theorem:main} adapted to Corollary~\ref{corollary:patch}, we know that $\rho = \mu \gamma$, which concludes the proof. (Note that  with $\gamma> \tilde{\gamma}(K)$ we will increase the error, while decreasing $\rho(\gamma) \le \tilde{\rho} = \mu \tilde{\gamma}$. This is why this choice is not sensible according to our theory.)
\end{proof} 
\end{corollary} 

\begin{corollary}[Patch-ups]
\label{corollary:patch}
Using Eq.\eqref{eq:alpha-tilde-bound} instead of Eq.~\eqref{eq:alpha-bound}, but then setting $\epsilon_i = 0$ yields the same results as before with the following changes: \\[3mm]
(a) Lemma \ref{lemma:lyapunov}: the bound becomes $\gamma \le \frac 1L \min \{ \frac 14 \frac{K \sigma}{K + c\sigma}, \frac {1-\sigma}{2} \} <  \frac{1}{6L}$. \\
(b) Theorem  \ref{theorem:main}: still using $\gamma = \frac{a}{4L}$, we require $a < \frac{2}{3}$ and we get a similar (slightly smaller) expression for $\rho$ as well as for the optimal step size $\tilde{\gamma}(K) := \frac{\tilde{a}(K)}{4L}$ replacing $\gamma^*$ in the theorem:
\begin{align} \label{eq:atilde}
\rho = \frac qn \frac{1-\frac 32 a}{1-\frac 12 a} \quad \text{and} \quad \tilde{a}(K) = \frac{2K}{1 + \frac3 2 K + \sqrt{1 + K + \left(\frac{3}{2}K \right)^2}} \geq \frac{2}{3} a^*(K) \, .
\end{align}
(c) The optimal asymptotic rate is  still $\tilde{\rho} \stackrel {n \to \infty} \longrightarrow  \frac qn$.
\begin{proof}
Redoing all proofs with an additional factor of $2$ on the RHS of  Eq.~\eqref{eq:alpha-bound}. One can also readily verify that the ratio $\frac{\tilde{a}(K)}{a^*(K)}$ (with $a^*(K)$ defined in Eq.~\eqref{eq:gammalk}) is a decreasing function of $K$, with value $1$ for $K=0$, and limiting value $\frac{2}{3}$ for $K \to \infty$.
\end{proof}
\end{corollary}

\subsection{Implementation details}

\paragraph{Construction of the neighborhoods required by q-SAGA and ${\cal N}$-SAGA:} 
For each datapoint~$i$, we want to define a neighborhood ${\cal N}_i$ (defined as the set of children of $i$ in a directed graph) such that for $j \in {\cal N}_i$,
$\| \alpha_j - \beta_j \|^2 \le \epsilon$. The approximation bounds in Section~\ref{sect:sharing} show that this distance is a function of $w$, which is not known {\it a priori}. In order to address this issue, we used the distance between data points $\delta_{ij} := \| x_i - x_j \|$ as a surrogate. The construction of the neighborhoods then amounts to constructing a directed graph on $n$ nodes by setting the $q$ nearest points to $j$ as its \emph{parents}.\footnote{This can be naively implemented by computing all pairwise distances $\delta_{ij}$ between data points ($O(n^2)$), but more efficient data structures using hashing~\cite{andoni2018LSH} or randomized partition trees~\cite{dasgupat2015RPT} can be used.} This ensures that $| \{i: j \in {\cal N}_i\}|=q$ $(\forall j)$, i.e. every $j$ has exactly $q$ parents.
Note that this simple construction can yield asymmetric neighborhoods (i.e. $j \in {\cal N}_i \centernot\implies i \in {\cal N}_j$); ${\cal N}_i$ is the set of children of $i$, and does not have to be of size $q$. 
One could also construct a symmetric neighborhood by defining $j$ to be a child of $i$ if their distance is less than $\sqrt{\epsilon}$ (which is a symmetric relationship), where $\epsilon$ is a constant chosen such that $q \approx 20$. In practice, we did not find this construction to yield better performance (in addition to violating the uniform $q$-memorization property). Note also that the above constructions ensure that $i \in {\cal N}_i$.

\paragraph{Growing $n$ heuristic:} For all the $q$-memorization algorithms, we used the same initialization heuristic proposed in~\cite{schmidt2013minimizing,defazio2014} for which during the first pass, datapoints are introduced one by-one, with averages computed in terms of the number datapoints processed so far (i.e. the normalization for $\bar{\alpha}$ is the number of different points seen so far instead of $n$).

\end{document}